\documentclass[pdflatex,sn-mathphys-num]{sn-jnl}
\usepackage[english]{babel}
\usepackage{geometry}
\usepackage{amsmath}
\usepackage{graphicx}
\usepackage{svg}
\usepackage{hyperref}
\usepackage[most]{tcolorbox}
\usepackage{xcolor}
\usepackage{adjustbox}
\newtcolorbox{promptbox}[1][]{
  enhanced,
  breakable,
  colback=gray!5,
  colframe=black!60,
  boxrule=0.6pt,
  arc=2mm,
  left=2mm,
  right=2mm,
  top=1mm,
  bottom=1mm,
  fonttitle=\bfseries,
  title=#1
}

\author[1]{\fnm{Alain} \sur{Gutierrez}}\email{alain.gutierrez@lirmm.fr}
\author*[1]{\fnm{Marianne} \sur{Huchard}}\email{marianne.huchard@lirmm.fr}
\author[2,3]{\fnm{Pierre} \sur{Martin}}\email{pierre.martin@cirad.fr}
\author[4]{\fnm{André} \sur{Miralles}}\email{andre.miralles@teledetection.fr}
\author[1]{\fnm{Violaine} \sur{Prince}}\email{violaine.prince@lirmm.fr}
\equalcont{The authors contributed equally to this work.}

\affil[1]{\orgdiv{LIRMM}, \orgname{Univ. Montpellier, CNRS}, \city{Montpellier}, \country{France}}
\affil[2]{\orgdiv{CIRAD}, \orgname{UPR AIDA}, \orgaddress{\street{F-34398}, \city{Montpellier}, \country{France}}}
\affil[3]{\orgdiv{AIDA}, \orgname{CIRAD, Univ. Montpellier},  \city{Montpellier},  \country{France}}
\affil[4]{\orgdiv{INRAE - UMR TETIS - Territoires, Environnement}, \country{France}}

\begin{document}

\title[Concept naming in FCA and RCA]{
A Variability-Based Framework for Interpretable Naming in Formal and Relational Concept Analysis}

\abstract{
Knowledge extraction from symbolic data often produces abstractions that are formally defined but not immediately interpretable by users. Formal Concept Analysis (FCA) and Relational Concept Analysis (RCA) provide representative settings for this issue: they generate explicit conceptual structures, implications, and relational dependencies from object descriptions and relations. Although these structures are explainable by design, their concepts are often identified by technical labels, which limits their use as human-interpretable knowledge units. Assigning meaningful names to such concepts is therefore a key issue for interpretation, navigation, validation, and reuse by domain experts.
This paper investigates concept naming in FCA and RCA from a symbolic knowledge representation perspective. We first characterize the linguistic and terminological challenges involved in naming generated symbolic abstractions, including ambiguity, discrimination, concision, and consistency across related concepts. We then propose a configurable framework for LLM-assisted concept naming. The framework relies on a variability model that controls which sources of information are exposed during naming, such as intent, extent, inherited information, neighboring concepts, implications, and relational attributes. It thereby makes explicit the semantic choices involved in moving from formal concept descriptions to human-readable names.
The approach is illustrated as a proof of concept on a small relational dataset in the pizzeria domain. This illustration shows how different configurations influence the names suggested by an LLM, and how naming variability can reveal interpretation choices, relational dependencies, and possible modeling issues in the underlying symbolic data.}


\maketitle


\section{Introduction}
\label{sec_introduction}
Knowledge extraction from symbolic data often produces abstractions that are formally defined but not immediately interpretable by users. These abstractions may take the form of concepts, rules, patterns, types, roles, or relational structures, whose meaning must be reconstructed from symbolic descriptors, e.g.  objects, attributes, relations, and structural context. A central challenge is therefore to attach semantic labels to such generated knowledge units, so that they can be inspected, compared, validated, and reused by domain experts. This challenge is related to semantic labeling and semantic table interpretation, where semantic labels are assigned to raw data elements using ontology classes, properties, or knowledge-graph entities \cite{DBLP:conf/esws/RamnandanMKS15,LIU2023100761}.

Among symbolic knowledge extraction approaches, Formal Concept Analysis (FCA) \cite{GanterFormalConceptAnalysis1999} and Relational Concept Analysis (RCA) \cite{HaceneRelationalconceptanalysis2013} provide a particularly suitable setting for studying this labeling problem. FCA offers a sound environment to structure and explore data represented as a set of objects described by a Boolean relation to identified attributes. It produces formal concepts, each grouping objects that share common attributes, and organizes them by a partial order, typically a concept lattice. FCA also supports the extraction of implications, so that conceptual structures and rule-like knowledge patterns can be explored together. 

While machine-learning approaches are often used to learn predictive models from data \cite{hastie2009elements,LeCun2015}, FCA and RCA are primarily aimed at structuring symbolic descriptions into explicit and inspectable knowledge units. This makes them complementary to machine learning: rather than producing a predictive model to be explained afterwards, they generate conceptual structures whose descriptions are available by construction. Multi-relational data are handled by RCA, a dedicated extension of FCA in which relational dependencies are included in concept descriptions. Both approaches are explainable by design, since each generated concept can be traced back to its extent, intent, and, in RCA, relational attributes. 

Yet, when FCA or RCA tools build such structures, concepts are often labeled only by meaningless identifiers, as in LatViz \cite{DBLP:conf/cla/AlamLN16} and FCA4J \cite{DBLP:conf/cla/GutierrezH022}, or are sometimes not labeled at all, as in ConExp\footnote{See \url{https://conexp.sourceforge.net/relatedprojects.html} and \cite{Hanika_Conexp-Clj_-_A_2019}}. 
In this setting, the general problem of semantic labeling becomes the more specific problem of assigning meaningful names to formal and relational concepts.
Recent advances in large language models (LLMs) offer a new opportunity to assist this task, provided that their suggestions are guided by explicit symbolic descriptions and validated by domain experts.

Early exploratory work has been proposed in this direction. Aranda et al. introduce an approach in which new attributes are added to name concepts that do not have proper attributes \cite{arandaConcepts2024}. Their approach is built around a single strategy and does not address the RCA setting. Guenoune et al., in turn, propose a naming approach for the specific case of extracting new classes in a conceptual model (UML) using RCA \cite{DBLP:conf/concepts/GuenouneGHLMMZ25}. These two approaches provide a basis for the present work, which aims to integrate them into a more systematic \textbf{framework} that accounts for both the complexity of formal and relational concept structures and the variability of possible naming strategies.

Indeed, the problem of concept naming is broad and useful in a number of FCA/RCA applications where concepts are handled as interpretable units. In conceptual modeling, the building of ontologies or conceptual models (e.g. UML-like class diagrams) involves different tasks, such as deriving candidate abstractions, factorizing shared properties, supporting ontology refinement, and guiding the merging, as presented in various works \cite{DBLP:conf/ijcai/StummeM01,DBLP:conf/iccs/GanterS03,DBLP:conf/ekaw/BendaoudNT08,DBLP:conf/iccs/HaceneVN11,DBLP:conf/ismis/MonninLNC17}. In this context,  a \textbf{concept} is interpreted as a \textit{candidate entity} (e.g. ontology concept, class, role) while the lattice structure supports the entity organization (e.g. subsumption). The \textbf{concept name} is therefore intended to become a \textit{meaningful identifier} of the modeled entity that remains stable across diagrams, documentation, and stakeholder discussions. Unlike  automatically assigned formal identifiers (character strings serving as keys), names are meant to give human users access to the meaning and role (e.g., location, cause) of the concept. Thus, relevant names help bridge  the gap between the structure and the domain language, making the extracted conceptualization easier to validate, communicate, and maintain. 

Moreover, during a visual navigation, e.g. in an exploratory search, a concept acts as a \textit{view} or a theme enabling the user to locate a node in a space portion, and get its meaning \cite{DBLP:conf/icfca/EklundDB04,DBLP:books/daglib/0014402,DBLP:conf/diagrams/Pattison14,Crampes2014,DBLP:conf/iccs/AndrewsH16,Alam2016,GreeneVisualizingexploringsoftware2017,GreeneConceptBasedExplorationRich,DBLP:journals/ijar/HuchardMMPRS24}. An appropriate and semantically informative concept name is thus crucial for an effective lattice navigation experience. Finally, in an information retrieval task, FCA offers an original form of exploration, allowing additional constraints, explicit goals, and interactive queries to be taken into account  \cite{Carpineto1996,S0943-7444(24)01202-6,DBLP:conf/icfca/CigarranGPV04,10.3217/jucs-010-08-0985,DBLP:journals/amai/CodocedoLN14}. Concepts organize results (documents, items, objects) as groups or facets that support iterative browsing. 

Providing a consistent name to concepts is subject to specific requirements depending on the FCA or RCA application. In conceptual modeling, an unclear or too general name can obscure meaning, blur distinctions between nearby concepts, and weaken the quality of the resulting ontology/conceptual model. Using a name made of various words may then clarify the meaning. Conversely, for visual navigation and information retrieval, a concept name should not be too long. It also has to be discriminant, and actionable, so that the user can quickly understand the clustering, compare alternatives, and reformulate its query. Determining meaningful names is therefore crucial to ensure their interpretability and usability. 

The naming process is challenging because a concept name must satisfy expectations of semantic and syntactic clarity.
The name has to follow an acceptable noun phrase syntactic pattern \cite{DBLP:conf/iwpc/FalleriHLNPD10}. Its length is thus measured in the number of atomic part-of-speech units necessary to ensure a non-ambiguous and discriminant semantic label. Naming depends on the data used to determine it, since various information from the lattice could be used to name a concept computed by FCA. 
This information may derive from the content of the concept itself, namely some or all of its attributes and objects, as well as from the content of related concepts. 
In the same way, the use of the names of more general or more specific related concepts  broadens or narrows the semantic domain of the concept name. 
Using information from the concept siblings  aims at defining the  concept semantic domain borders (i.e. discriminant information), that may constitute a negative constraint on the  concept naming possibilities, to enhance the terminological relevance. 
When naming concepts computed by 
RCA, the information listed above but associated with relational dependencies can also be added. Therefore, the  adopted information is a source of \textbf{variability} in the naming. In an extreme situation where the concept to be named contains no proper objects or attributes, the semantic variation of its possible names can be significant as its names can exclusively be provided by related concepts. Finally, when all concepts in a set should be named using the name of related concepts, the order of naming has to be defined, knowing that starting at the top of the lattice, bottom of the lattice, or any other concept may affect the names and thus increase the variability. 

This work investigates the challenge of naming concepts in structures produced by FCA and RCA using an LLM-assisted approach. The LLM acts as a terminologically aware helper, and is assigned the following goal:  to suggest names to concepts, using the information contained 
in the prompt. 
Different prompt configurations can then be derived from the selected sources of formal information, making it possible to examine how these sources influence the naming process.
Thus, the aim of this paper is to: (i) characterize concept naming as a semantic labeling problem for generated symbolic abstractions, with specific linguistic and terminological challenges; (ii) propose a variability-based framework that makes explicit which sources of formal information are used to guide naming; and (iii) illustrate, as a proof of concept, how different configurations influence LLM-suggested names on a small relational dataset. 
We have deliberately chosen a small dataset in order to avoid  the possible disturbances that may arise from large datasets (e.g. difficult readability, possible duplicates or near duplicates, technical complexity in applying RCA, etc.). Scope and volume could be an issue per se. We first wanted to tune the process on a controllable amount of elements to be named before extending it on a larger number of  conceptual nodes as well as relations. The chosen dataset is applied to a specific scenario —-buying a pizza-— which does not require any specialized expertise. Moreover, this dataset illustrates the main issues raised by concept naming.

Section \ref{sec_basics} introduces FCA, RCA, and the small RCA dataset. Section \ref{sec_problemFraming} outlines the issues and the challenges for naming according to linguistic principles. 
Section  \ref{sec_naming} presents the variability-based framework, including naming guidance principles, the formal language used to express variability (UVL), the main sources of variability and their modeling in UVL, and the operational pipeline used to systematically generate prompt variants according to selected preferences.
Section \ref{sec_evaluation} illustrates the framework on the dataset and analyzes the resulting naming variability.
Section \ref{sec_related} describes related work. Section \ref{sec_conclusion} concludes and presents perspectives of this work.


\section{FCA and RCA in a Nutshell}
\label{sec_basics}

\paragraph{Formal Concept Analysis (FCA).}
\label{sec_fca}
Formal Concept Analysis (FCA) provides a sound framework, based on lattice theory, to organize and explore data given in a formal context (FC) 
${\cal K}=(G,M,I)$, where $G$ and $M$ denote the set of objects and the set of attributes, respectively, and the incidence relation $I \subseteq G\times M$ specifies which objects have which attributes~\cite{GanterFormalConceptAnalysis1999}. Our test dataset is provided in Table \ref{table_pizzas}.  
This table shows FC Ingredient (on the top right) which describes some ingredients used in pizza preparation (e.g. goatcheese, burrata) based on their characteristics related to the seasons in which they are suitable to eat (spring, summer, autumn) and feeding behaviors (vegetarian, vegan). 

\begin{figure}[htb]
\centering
  \includegraphics[width=\textwidth]{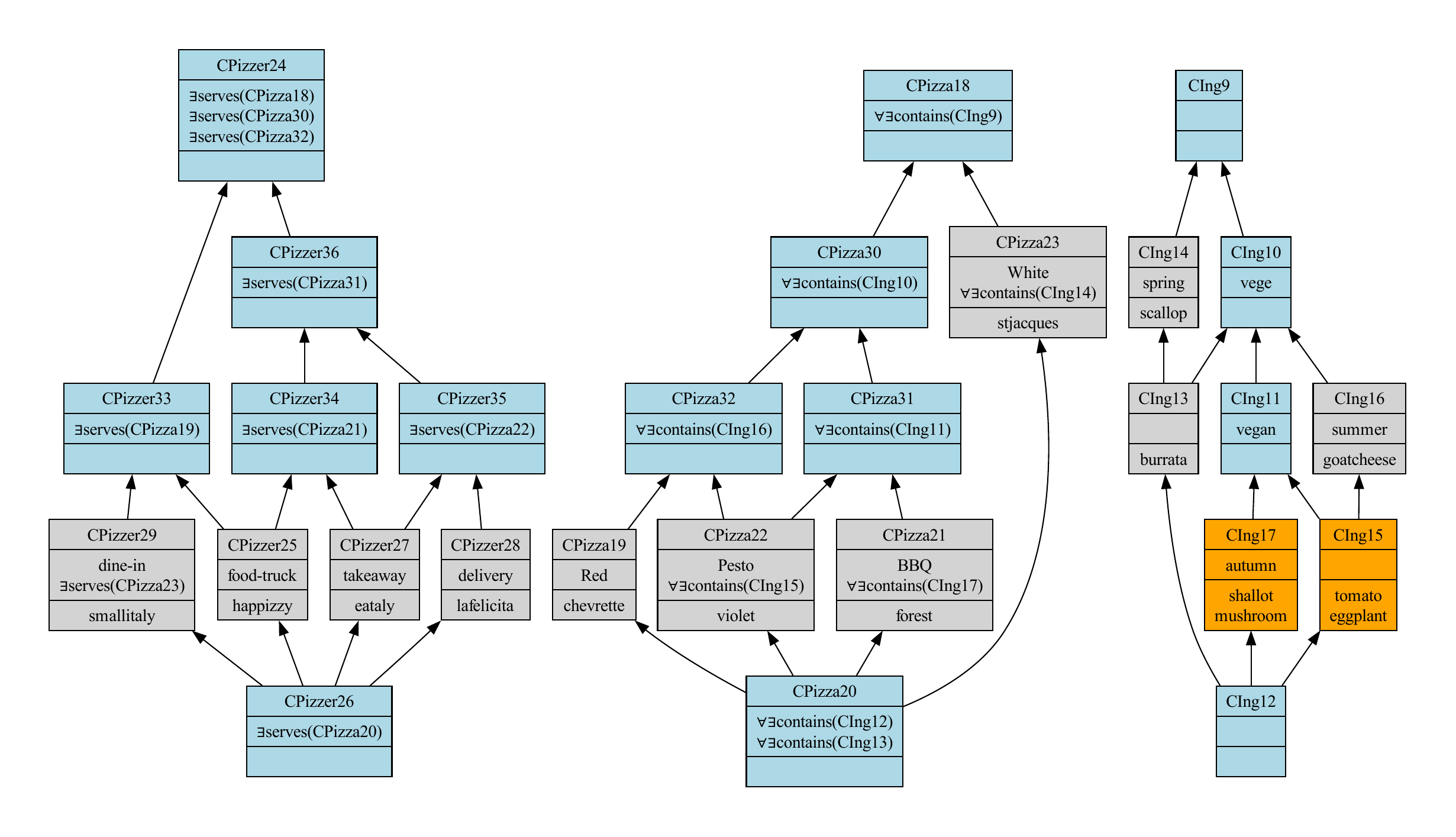}
  \caption{Concept lattices of FC pizzerias (left), pizzas (middle), and ingredients (right) with relational attributes, at the end of the RCA computation process. A lattice node represents a concept consisting of three sections that indicate, from top to bottom, the concept name, its intent, and its extent, both reduced to proper attributes and objects respectively. In this figure, the concept name is a formal identifier and is the element to be replaced by the naming process. The lattices are built using FCA4J~\cite{DBLP:conf/cla/GutierrezH022} and wording was reduced.
  }
  \label{fig_pizzanora}
\end{figure}

\begin{table*}[htb]
\centering
\caption{The small RCA Dataset represented as a relational context family adapted from \cite{DBLP:conf/concepts/GutierrezHMZ25}. Pizzerias serve pizzas that contain ingredients. The three formal contexts and the two relational contexts are presented at the top and bottom respectively. 
}
\label{table_pizzas}
{\footnotesize
\setlength{\tabcolsep}{3pt}
\begin{tabular}{|c|c|c|c|c|}
\hline
Pizzeria&
\rotatebox{70}{food-truck}&
\rotatebox{70}{takeaway}&
\rotatebox{70}{delivery}&
\rotatebox{70}{dine-in}\\\hline
happizzy&$\times$&&&\\\hline
eataly&&$\times$&&\\\hline
lafelicita&&&$\times$&\\\hline
smallitaly&&&&$\times$\\\hline
\end{tabular}
\begin{tabular}{|c|c|c|c|c|}
\hline
Pizza&
\rotatebox{70}{Red}&
\rotatebox{70}{BBQ}&
\rotatebox{70}{Pesto}&
\rotatebox{70}{White}\\\hline
chevrette&$\times$&&&\\\hline
forest&&$\times$&&\\\hline
violet&&&$\times$&\\\hline
stjacques&&&&$\times$\\\hline
\end{tabular}
\begin{tabular}{|c|c|c|c|c|c|}
\hline
Ingredient&
\rotatebox{70}{vege}&
\rotatebox{70}{vegan}&
\rotatebox{70}{spring}&
\rotatebox{70}{summer}&
\rotatebox{70}{autumn}\\\hline
goatcheese&$\times$&&&$\times$&\\\hline
burrata&$\times$&&$\times$&&\\\hline
scallop&&&$\times$&&\\\hline
tomato&$\times$&$\times$&&$\times$&\\\hline
shallot&$\times$&$\times$&&&$\times$\\\hline
mushroom&$\times$&$\times$&&&$\times$\\\hline
eggplant&$\times$&$\times$&&$\times$&\\\hline
\end{tabular}
}
{\footnotesize
\setlength{\tabcolsep}{3pt}
\begin{tabular}{|c|c|c|c|c|c|c|c|}
\hline
serves&
\rotatebox{70}{chevrette}&
\rotatebox{70}{forest}&
\rotatebox{70}{violet}&
\rotatebox{70}{stjacques}
\\\hline
happizzy &x&x&& \\\hline
eataly &&x&x& \\\hline
lafelicita &&&x& \\\hline
smallitaly &x&&&x \\\hline
\end{tabular}
\begin{tabular}{|c|c|c|c|c|c|c|c|}
\hline
contains&
\rotatebox{70}{goatcheese}&
\rotatebox{70}{burrata}&
\rotatebox{70}{scallop}&
\rotatebox{70}{tomato}&
\rotatebox{70}{shallot}&
\rotatebox{70}{mushroom}&
\rotatebox{70}{eggplant}
\\\hline
chevrette &x&&&x&&& \\\hline
forest &&&&&x&x&\\\hline
violet &&&&x&&&x\\\hline
stjacques &&x&x&&&&\\\hline
\end{tabular}
}
\end{table*}
%
Two main functions respectively associate a set of objects with their shared attributes ($\alpha:\mathcal{P}(G)\rightarrow \mathcal{P}(M)$, such that $\forall X \subseteq G$ $\alpha(X)=\{y\in M|\forall x \in X, (x,y) \in I\}$) and a set of attributes with their shared objects 
($\beta:\mathcal{P}(M)\rightarrow \mathcal{P}(G)$, such that $\forall Y \subseteq M$, $\beta(Y)=\{x\in G| \forall y \in Y, (x,y) \in I\}$). 
A concept is a pair $(X,Y)$, where $X \subseteq G$, $Y \subseteq M$, and  $\alpha(X)=Y$ (which is equivalent to  $\beta(Y)=X$).
$X$ and $Y$ are respectively named the intent and the extent of the concept. 
For FC Ingredients, 
\texttt{(\{shallot, mushroom, tomato,eggplant\},\{vege,vegan\})} is a concept labeled \texttt{CIng11} in Figure \ref{fig_pizzanora} (right-hand side).

\begin{figure}[htb]
\centering
  \includegraphics[width=\textwidth]{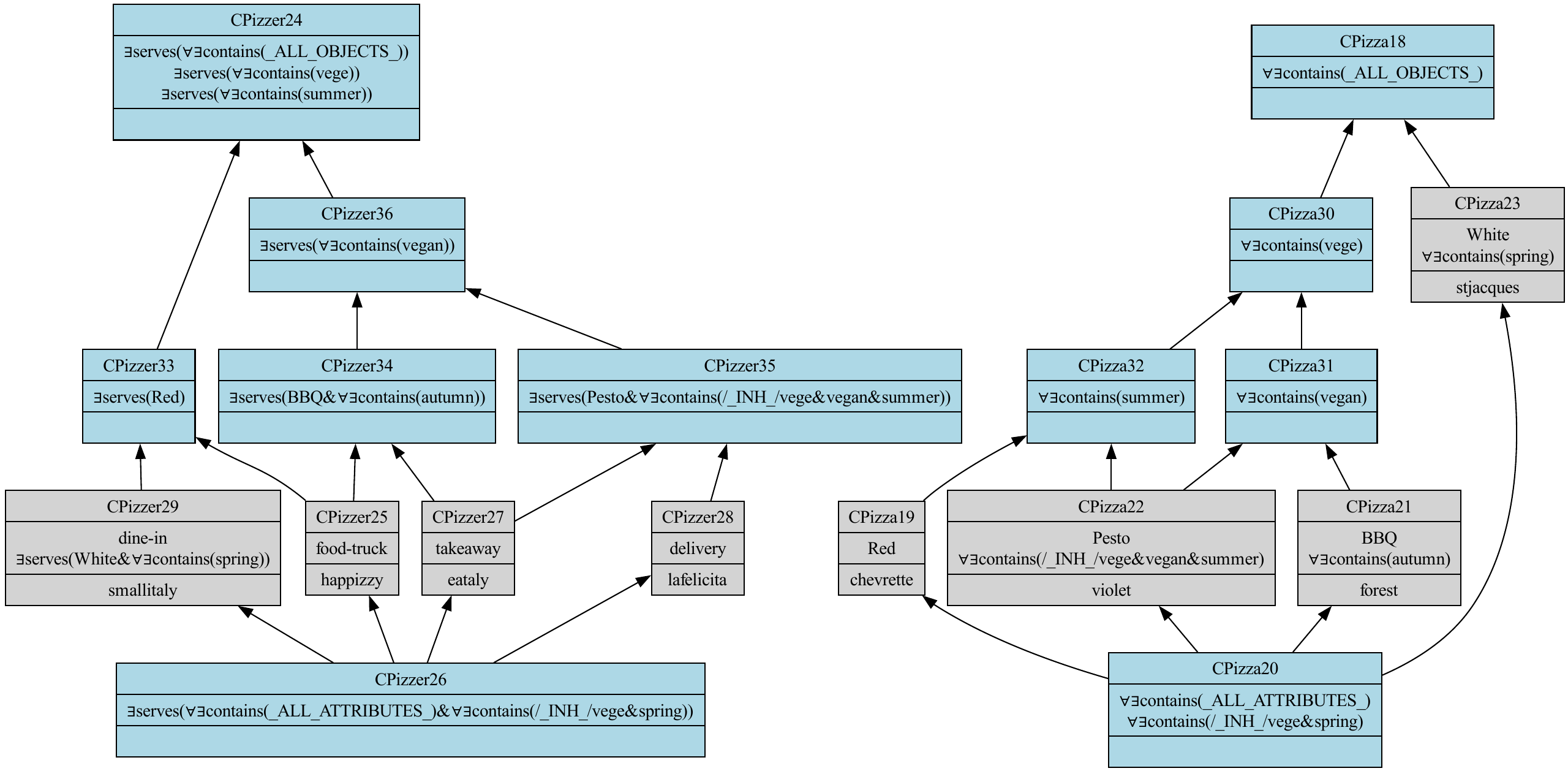}
  \caption{Concept lattices of FC pizzerias (left) and pizzas (right) with \textbf{developed} relational attributes, at the end of the RCA computation process. The lattices are built using FCA4J~\cite{DBLP:conf/cla/GutierrezH022} and wording was reduced. 
  }
  \label{fig_pizzarai}
\end{figure}

The concept lattice ${\cal L(\mathcal{K})}$ computed on top of $\mathcal{K}$, is the set of concepts  provided with a partial order relation based on inclusion of concept extents: a concept $(X_{sub},Y_{sub})$ is a sub-concept of a concept $(X_{sup},Y_{sup})$ when $X_{sub} \subseteq X_{sup}$. 
The concept lattice associated with FC ingredient is shown in the right-hand side of Figure \ref{fig_pizzanora}.
In Figures \ref{fig_pizzanora} and \ref{fig_pizzarai}, 
concepts are displayed in a simplified form, where only proper or  introduced\footnote{We use the terms ``proper'' or ``introduced'' instead of ``generating''. ``proper'' highlights the semantics, while ``introduced'' emphasizes the position of the attribute or object in the concept lattice. ``generating'' is another term of the literature which emphasizes the fact that from one of these objects (resp. attributes), the concept can be generated by application of the closure operator $\beta \circ \alpha$ (resp. $\alpha \circ \beta$).}
objects and attributes are shown. For instance \texttt{CIng11} is a sub-concept of \texttt{CIng10}. Its extent is composed of bottom-up inherited objects: $\{shallot, mushroom, tomato, eggplant\}$. It has no introduced objects.  $shallot$ and $mushroom$ are introduced in \texttt{CIng17}, while $tomato$ and $eggplant$ are introduced in \texttt{CIng15}.
\texttt{CIng11} intent is $\{vege, vegan\}$, with $vegan$ introduced in \texttt{CIng11} and $vege$ introduced in \texttt{CIng10}.

The FCA framework also provides implication such as $Pr\rightarrow Co$, where $Pr \subseteq M$ and $Co \subseteq M$ that indicates that all objects owning the attributes in the premise $Pr$ also own those in the conclusion $Co$. $\{autumn\} \rightarrow \{vegan\}$ is an example of implication in the FC Ingredient. It is held by $shallot$ and $mushroom$.

\paragraph{Relational Concept Analysis (RCA).}
\label{sec_rca} 
RCA is an extension of FCA to multi-relational datasets based on entity-relationship models~\cite{HaceneRelationalconceptanalysis2013}. 
The data are represented in a \textit{relational context family} (RCF), which is a pair $(\mathcal{C},\mathcal{R})$.
$\mathcal{C} = \{\mathcal C_i = (\mathcal G_i, \mathcal M_i,  \mathcal I_i)\}$ is a set of FC. 
The attributes of any $\mathcal M_i$ are called primitive attributes.  
$\mathcal{R} = \{r\ |\ r \subseteq \mathcal G_i \times \mathcal G_j\}$ is a set of relations $r$ (relational contexts RC) from objects of a \textit{source context} $src(r)=\mathcal C_i$ to objects of a \textit{target context} $dest(r)=\mathcal C_j$.  $r(o)$ denotes the \textit{image} of $o\in \mathcal G_i$ by relation $r$.
Table \ref{table_pizzas} presents an RCF  composed of three FC and two RC, corresponding respectively to entities (pizzerias, pizzas, ingredients) and relationships (serves, contains). 
To consider relationships in the concept formation, RCA introduces \textit{relational attributes} that gradually extend FCs.
A relational attribute has the form $qr(C)$, where $q$ is a quantifier, $r$ a relation and $C$ a concept.
An example of relational attribute is $\forall\exists$ \texttt{contains(CIng11)}, which means ``contains at least one and only ingredients of \texttt{CIng11} extent'', i.e. vegan ingredients.
This relational attribute can be assigned to pizzas \texttt{violet} and \texttt{forest} that respectively contain \texttt{tomato, eggplant} and \texttt{shallot, mushroom}. Notice that although these pizzas initially share no primitive description — and do not even share any ingredient via the \texttt{contains} relation — the elicitation of this relational attribute provides them a shared property.
During the complete RCA process on an RCF $(\mathcal C,\mathcal R)$, each FC 
$\mathcal C_i = (\mathcal G_i, \mathcal M_i, \mathcal I_i)$ of $\mathcal C$ is extended using the set $R$ of relations in $\mathcal R$, for which $\mathcal C_i$ is the source, and a set of  scaling operators $\mathcal Q$  (i.e. quantifiers, e.g. $\exists$ or $\forall\exists$) associated with the relations of $R$, via a mapping $\rho : \mathcal{R} \rightarrow \mathcal {Q}$. 
The extended FC is $\mathcal C_i^+ = (\mathcal G_i, \mathcal M_i^+, \mathcal I_i^+)$, which enriches  $C_i$ with relational attributes  as follows: 
$\mathcal M_i^+ = \mathcal M_i \cup \{qr(C)\ |\ q\in \mathcal Q, r\in R, 
\rho(r)=q, ~C \in \mathcal L(dest(r))\}
$, ~and~$\mathcal I_i^+ = \mathcal I_i \cup \{(o,qr(C))\ |\ C = (E,I)$ ~and $~conforms(q,E,r(o))\}$.
$conforms(\exists,E,r(o))$ is defined by $E \cap r(o) \not = \emptyset$, i.e.  $o$ is connected by $r$ to at least one element in $E$.
$conforms(\forall\exists,E,r(o))$ is defined by $r(o) \not = \emptyset$, and $r(o) \subseteq E$,  i.e. $o$ is connected by $r$ to at least one element in $E$ and only to elements in $E$. 
The set of all extended FC is $\mathcal C^+ = \{\mathcal C_i^+\ |\ \mathcal C_i\in \mathcal C\}$.
The RCA process then iterates until a fix-point or once the number of steps defined by the user is reached.  
At step $s, s \in \{0, .. n\}$, RCA builds (1) the concept lattices of FCs in the RCF (extended when $s \geq 1$), and (2) a new RCF $(\mathcal C^+, \mathcal R)$ where the relational attributes are built using all known  concepts.
Figure \ref{fig_pizzanora} shows the concept lattices at the fix-point.
\texttt{CPizza31} groups the pizzas $violet$ and \textit{forest}, which share $\forall\exists$ \texttt{contains(CIng11)} (i.e. they contain only vegan ingredients). 
After discovering \texttt{CPizza31}, RCA was able to group the pizzerias $happizy$, $eataly$, and $lafelicita$ in \texttt{CPizzer36}, since they share the attribute $\exists$ \texttt{contains(CPizza31)}.
Interestingly, the extended FC $\mathcal C_i^+$ built at each step can be used to compute implications. For example, $\{\exists$ \texttt{serves(CPizza22)}$\} \rightarrow \{\exists$ \texttt{serves(CPizza31)}$\}$ can be computed from the last extended FC Pizzerias.

It becomes quite clear that interpreting a concept 
is difficult, particularly when it contains relational attributes whose meaning is obscure (e.g. $\forall\exists$ \texttt{contains(CIng11)}). This has led to implement rewriting approaches that rely on an expanded form of relational attributes \cite{DBLP:journals/ijgs/DolquesBHG16,DBLP:conf/icfca/DolquesBHB19,DBLP:conf/kcap/WajnbergVLPM19,DBLP:conf/concepts/MusslinBHMPRS24,DBLP:conf/concepts/GutierrezHMZ25}. In this section, the used rewriting, named \texttt{Rai}, standing for ``\textbf{r}ewriting using proper \textbf{a}ttributes, alternatively use \textbf{i}nherited'', consisted in replacing each concept with the attributes introduced by its intent, or inherited when no attribute is introduced \cite{DBLP:conf/concepts/GutierrezHMZ25}. Figure \ref{fig_pizzarai} presents the concept lattices of FC pizzerias and pizzas with rewritten relational attributes.
For instance, in concept CPizzer36, the relational attribute \texttt{$\exists$serves(CPizza31)} is first developed as 
\texttt{$\exists$serves($\forall\exists$contains(CIng11)} and then as
\texttt{$\exists$serves($\forall\exists$contains(vegan))}.
In concept CPizzer35, the relational attribute \texttt{$\exists$serves(CPizza22)} is first developed as 
\texttt{$\exists$serves(Pesto \& $\forall\exists$contains(CIng15)} and then as
\texttt{$\exists$serves(Pesto \&   $\forall\exists$ contains (vege \& vegan \& summer))}.
As concept \texttt{CIng15} does not introduce any attributes, its inherited attributes are used and are preceded by “/\_INH\_/”.


\section{Issues and challenges for naming}
\label{sec_problemFraming}

Producing a symbolic description belongs to the terminology field. This section presents the linguistic complexities in terminology which burden the naming process.
In our context, naming a concept means attributing the most appropriate symbolic description that respects the constraints of its final use, e.g. it has to be meaningful, syntactically correct, and short (length measured in the number of atomic part-of-speech units) for visual navigation.  
First, we discuss the overall issues of terminology (Sect.~\ref{sec_terminology}). 
Then, we review its consideration using FCA (Sect.~\ref{sec_terminologyFCA}).
Finally, we identify concrete guidance principles for naming  (Sect.~\ref{sec_guidance}).

\subsection{Terminology}
\label{sec_terminology}

Natural language is subject to several communication side effects, among them redundancy and elision. \textbf{Redundancy} is when several discourse chunks address the same global meaning: their abundance is sometimes pedagogical but may lose the point and become imprecise or tedious. \textbf{Elision} is when the discourse chunk is the smallest pointer to a meaning, thus needing a complete set of information to restore the path to the intended meaning. 
These side effects exist at every level of the discourse, and especially at the terminological one, creating ambiguity and imprecision  caused by elision and redundancy respectively. 

In data modeling and conceptual design, ambiguity and imprecise lexical relations are a major source of noise: The same term may denote different notions depending on the context, while different terms may be used interchangeably for the same notion, leading to inconsistent annotations and unstable concept descriptions. This noise propagates to downstream analyses, e.g., it fragments otherwise coherent groups and inflates the number of near-duplicate concepts. 
To assess how this noise may impact the results, the next paragraphs of this section first present the main lexical relations related to the communicative aspect of language (i.e., synonymy, polysemy, homonymy, and metonymy) and then four knowledge representation relations used in modeling (i.e., hypernymy, hyponymy, antonymy, and meronymy).

\paragraph{Synonymy, the redundancy issue.}

\textbf{Synonymy}  can be defined \textit{per se}  as the equivalence relation. A term A is a synonym of B if it can be substituted to B without modifying the meaning or the value of what has been designated by B.
However, the equivalence relation cannot be satisfied in a context-free approach. In natural language, synonymy fails to be transitive, and has to be replaced by a context-based relation, what has been called a \textbf{relative synonymy}  \cite{DBLP:conf/nlprs/LafourcadeP01}.
An example of relative synonymy for ingredient description  
could be “vegetables / veggies" as well as “hot / spicy".
Examples for ingredient names are “shrimp / prawn" or “calamari / squid"; and for relations  
 in the pizzeria domain, it could be “hasIngredient / containsIngredient".

\paragraph{Polysemy, Homonymy and Metonymy, the elision issue.}

Polysemy is an issue by itself in the computational field that deals with terminology and several models, including FCA-based approaches \cite{DBLP:journals/amai/Priss22}, have tried to reduce its impact. 
\textbf{Polysemy} is defined as the ability to assign  different meanings to the same term. 
This notion covers a large domain of lexical relations according to the ties between the different meanings, and the reasoning path that relates a meaning to another. If these meanings are independent (i.e, there is no apparent path from meaning A to meaning B), then it is a pure \textbf{homonymy}. There is a coincidence in the character string representing completely separated notions. Example: table (the furniture) and table (the data structure).\\
Let us suppose that term T has at least two meanings, A and B.
The most common polysemous situation is when the path from meaning A to meaning B has the following features: A could be an attribute value of B, or A and B share a common partial description, role, or usage. 
It could be discriminated by the domain context, in which the scope of the context window becomes an asset. 
Examples: “A is an attribute value of B" such as the term “orange" describing ingredients, where  orange can correspond to a fruit category or a color (a fruit has a color so color could be derived from fruit category) in relation to the context; 
“A and B share common descriptions, roles or usage" such as the term “base" in the pizza domain, where base can correspond to the pizza crust or a sauce base (same role), and such as the  terms “pepper" (ground) and “peppercorn" in a description of ingredients as both are hot. \\
Last, a particular type of lexical relation, metonymy, creates a polysemous situation. \textbf{Metonymy} is a discourse  elliptic process that allows the substitution of a predicate argument by another, in order to shorten communication. The predicate is a verb, and in our framework, is represented by a relation between objects. A classical example of metonymy is when a location name is  substituted for someone located there. “London has called" refers to “our correspondent / somebody / an institutional in London has called". Here, there is an ambiguity about the agent. 
Metonymy creates an additional polysemous situation when the predicate arguments are not distinguished. 
An example for relations  
in the pizzeria domain could be the relation “serves" (a customer / a pizza). But the term “serve" is polysemous \textit{per se}. So in a given situation, even when polysemy could be resolved by context, if the relation  design is “flawed" by a metonymic process, then a polysemous situation may still remain.

\paragraph{Hypernymy/hyponymy, the symmetrical is-a ontological relation}

A term A is a \textbf{hypernym} of a term B if the meaning of A acts as a parent category of the meaning of B. 
Dictionary definitions are generally configured as sentences providing the closest genus (hypernym) and the most relevant species (specific features). Example from the Cambridge Dictionary for the term “pizza": \textit{a large circle of flat bread baked with cheese, tomatoes, and sometimes meat and vegetables spread on top}. The genus, or local hypernym is “a large circle of flat bread" which hypernym is a “flat bread", and its hypernym is “bread". The specific feature is “baked with cheese, tomatoes, and sometimes meat and vegetables spread on top". The Oxford English Dictionary  provides another parent category in its definition: \textit{A savoury dish of Italian origin, consisting of a flat, usually round base of dough, baked with a topping of tomatoes, cheese, and any of various other ingredients, such as meat, anchovies, or olives.} The latter definition indicates two possible parent categories, i.e., Dishes and Dough.  
In addition, if A is a hypernym of B, then B is a \textbf{hyponym} of A. 
This relation explains why the terms are presented in pairs.
Thus, if A is polysemous then there are several pairs with A as the left member.
If B is polysemous, then the specific meaning of B which makes B a hyponym of A has to be indexed by the context. 
An example for 
ingredients could be vegetarian / vegan.

\paragraph{Meronymy, the part-whole ontological relation}

\textbf{Meronymy}  is the part–whole relationship. A and B are in a meronymy relation if B is the term  designating a complex structure and A is the term designating a part of the B named structure. Unlike hypernymy/hyponymy and polysemy, A and B do not necessarily share a common description (e.g “car" and “wheel"). Unlike polysemy and homonymy, they may not share a common term (preceding example). Unlike synonymy, they do not share a common meaning. However, meronymy can be used in a metonymic process such as using the part instead of the whole. This will create a situated polysemy that is a liability in data modeling and design. Also, meronymy may associate terms that are polysemous (e.g.“car" and “wheel" or “pizza" and “crust", where  wheel and crust are polysemous). In that case, it is better to rename the part with a multiword expression including the structure name (car wheel or pizza crust). 
Thus, meronymy  is a structuring relation that can be exploited, but it may also introduce bias if mixed in without control.
 
\paragraph{Antonymy, inverse or complement? }
The term A is an \textbf{antonym} of B if A points at a meaning that could be seen as an opposite of a meaning of B. 
It should not be confused with negation, e.g. not vegan is not always meaty.
Antonymy is a scalable relation \cite{DBLP:conf/coling/SchwabLP02}. ``Hot'' is the antonym of ``cold'' but also of ``mild", depending on the utterance context. ``Mild" could be the antonym of both ``hot" and ``cold". 
Examples of antonym terms could be ``spicy" and ``mild". As one can see, ``mild" is polysemous, and ``spicy" has a synonym, ``hot", that has also  an  antonymic relation with mild. \\
Tricky situations could be observed. 
E.g. ``vegan" and ``vegetarian" may be modeled as related dietary categories, as, in many practical taxonomies, ``vegan" is treated as a subtype of the vegetarian/plant-based space (because vegan food also excludes meat/fish, and additionally excludes eggs/dairy and other animal products), using inclusion-like reading in food annotation settings.
However, in an application-specific context (e.g., a pizza menu, a faceted search UI, or a dataset with a single ``diet label'' field), ``vegan" and ``vegetarian" may behave as pragmatic opposites if the annotation scheme enforces mutually exclusive labels. For example, if each pizza must receive exactly one label among {vegan, vegetarian, meaty, seafood}, then vegan and vegetarian have become contrasting labels in practice, even though they are not strictly antonyms linguistically. In that case, the opposition comes from the classification design, not from the lexical relation. 

\subsection{How Terminology Shapes Formal Concept Analysis}
\label{sec_terminologyFCA}

Representing data provided in natural language for FCA implies expressing this data in terms of objects and attributes. For RCA, this also implies considering relations. The process of determining and naming objects, attributes, and relations is thus affected by lexical relations related to the communicative aspects of language and knowledge representation relations.

\paragraph{Synonymy may prevent relevant abstractions emergence}
Attribute synonymy (i.e., several attributes are  synonymous in the context) prevents proper factorization and the discovery of useful abstractions, because semantically equivalent (or near-equivalent) information is spread across different attributes. Moreover, since the domain of the context is not always precisely provided, then what could be seen as a synonym (e.g. an attribute “designation" seen as a synonym of “name") does not exactly cover the same meaning or the same domain. This approximate synonymy may lead to a possible confusion when manually updating factorization outcomes. For objects, a similar issue arises when different object names refer to the same entity (duplicates, aliases, naming variants) which creates  artificial distinctions in the extents. 
In addition, this entity can have its descriptions accidentally different (e.g. complementary).
In RCA, the problem extends to relations as well: a synonymous or inconsistently named relation (e.g.  hasIngredient vs containsIngredient) generates redundant equivalent relational attributes with the same consequence as for primitive attribute synonymy. 

\paragraph{Polysemy may merge unrelated descriptions/objects} 

Polysemy introduces a different kind of noise than synonymy: instead of splitting equivalent information, it will collapse distinct meanings into a single attribute/object/relation name. A polysemous attribute name mixes different properties under one label (e.g., pepper as a spice vs. a vegetable), which creates  spurious co-occurrences and incorrect factorizations and abstractions. This extends to relational attributes when relation names are polysemous or are used in a metonymic way. Polysemous object names, by causing the merge of different entities under the same object label, artificially combine extents and hide meaningful distinctions between concepts.
The particular case of homonymy typically yields more misleading concept groupings, which is reinforced by the propagation in RCA.

\paragraph{Hypernymy and hyponymy may give opportunities for abstraction / refinement}
Attribute hypernymy/hyponymy is often encoded in the formal context, with a simple strategy stating when an object owns the hyponym, then it owns the hypernym (e.g, if an object owns “a large circle of flat bread", then it owns “a flat bread"). This encoding, which can be considered as a scaling~\cite{GanterFormalConceptAnalysis1999}, has been proven to be useful for avoiding missing valuable abstraction in various situations \cite{DBLP:conf/oopsla/GodinM93}. Alternatively, attribute hypernymy/hyponymy taxonomies can be accounted for during the computation \cite{DBLP:journals/ijfcs/CellierFRD08}. An example of formal objects in a context describing ingredients could be cheese / mozzarella. In a consistent encoding, hypernym description (attributes) is included in hyponym description, except in rare cases (e.g. ostrich cannot fly, while flying has been assigned in general to bird, a hypernym of ostrich). An example of relations in a relational context family in the pizzeria domain could be hasIngredient (includes dough and topping) / hasTopping. Encoding this situation in RCA could be by reifying the relations in formal contexts or describing them at a higher-level, such as the encoding proposed for UML models \cite{DBLP:conf/concepts/GuenouneGHLMMZ25}. 

\paragraph{Meronymy may introduce ambiguity about inclusion levels}
Moreover, meronymy (part–whole relations) mainly affects the granularity of descriptions. 
Typical examples for formal attributes to a pizza could be crust, topping, and sauce. The main question is whether these attributes correspond to parts of the pizza. 
An example for formal objects in a context describing ingredients could be onion / slice. When substituting ``onion slice" to ``slice", then the meronymic relation between objects is no more ambiguous (since ``slice" could be assigned to several different ingredients, e.g. cheese, tomato, etc.). 
An example of relations for a relational context family in the pizzeria domain could be “prepare" viewed as ``makesDough", ``addTopping", ``addSauce", and ``bake".
Mixing whole-level and part-level terms in the same formal context may produce concept descriptions at heterogeneous levels of abstraction, which complicates interpretation and naming. In RCA, meronymy can be handled more explicitly through dedicated relations (e.g., hasPart, hasIngredient, hasTopping), which improves semantic clarity; however, if part–whole information remains implicit in lexical labels only, it can still introduce ambiguity and redundancy. 

\paragraph{Antonymy may introduce presence / absence ambiguity}
Finally, antonymy mainly raises a modeling issue rather than a purely lexical one. Because FCA uses binary attributes, the absence of an attribute does not necessarily imply its opposite,  especially in partial contexts. For example, not assigning the attribute “spicy” to an object does not imply that it is ``mild"; it may simply suggest missing information. For this reason, antonymic distinctions are often best modeled through explicit opposite attributes (e.g. spicy and mild, thin\_crust and thick\_crust) when the underlying dimension is well defined. When three or more attributes are opposites, e.g. hot / mild / cold, the modeling solution consists of scaling. The use of antonyms creates a clear contrast in concepts and implications. However, it must be applied selectively, as an attribute does not necessarily have a meaningful antonym, and excessive duplication can increase redundancy and reduce the lattice readability.
Negation, which, as we have seen, differs from antonymy, may be encoded in FCA using an attribute $a$ together with its negated counterpart 
$not~a$.
In RCA, the systematic use of antonymic relations (e.g. is served hot / is served cold) ensures consistency.

\paragraph{In practice, data are incomplete and heterogeneous, names are FC-dependent} Formal and relational contexts are generally incomplete. On the one hand, an attribute or its taxonomy may be missing. E.g. if the attributes of FC Pizzeria are dine-in / takeaway / delivery / food truck, then the three last attributes may be generalized by off-premise. This generalization may help in naming \texttt{CPizzer36}. On the other hand, missing attributes that are less important than others (e.g., pizza size) prevent objects from being separated (e.g., pizzas according to their format). Missing objects also induce erroneous data dependencies. Conceptual exploration \cite{DBLP:books/sp/GanterO16} can help supplement FC by presenting dependencies between data and asking experts to identify additional objects that would invalidate them.  
Moreover, FC generally contains different categories of information. For instance, FC Ingredient in Table \ref{table_pizzas} contains two categories of information, i.e. 
the season where it is suitable to eat an ingredient (autumn, spring) and feeding behavior (vegetarian, vegan). In the corresponding lattice (Figure \ref{fig_pizzarai}),  seasons are spread in  concepts located at diverse levels of the graph rather than at the same level, making it  difficult to identify them as different “values" of seasons. A solution to prevent such misunderstanding may be to insert the category in the name of the attributes, e.g., season:spring and season:autumn. When applied to a multi-level taxonomy, the naming must take into account each of the levels. This multiple formulation can reduce the dependency of attribute names on FC. Similar considerations must be taken into account when naming objects and relations. 

\subsection{Guidance for naming concepts in FCA and RCA}
\label{sec_guidance}

Beyond the review of the terminology presented in Sect. \ref{sec_terminology}, three fundamental principles may be proposed to guide the naming of concepts.

\paragraph{Identify primary and secondary attributes}

To build the concepts, FCA treats all attributes uniformly from a formal standpoint. However, for their interpretation and therefore their naming, not all the attributes contribute equally: some are core of the FC domain, meaning that they are primary in the intended meaning of the concept, while others are peripheral or secondary as they may be less informative or redundant. 
For instance, in a context describing ingredients with attributes from the diet categories (e.g., vegan, vegetarian) and seasons of availability (e.g., summer, spring), a concept with intent \{vegan, summer\} (such as \texttt{CIng15} in Fig.~\ref{fig_pizzanora}) should be named primarily through the diet attribute (vegan) as this information is timeless, while the seasonal attribute (summer) is more a situational factor (because availability is occasional) that may serve as refinement. 
Seasonal attributes are indeed less important naming cues in relation to the FC domain than those in the dietary category.
When applied to RCA, this prioritization includes the relational attributes.

\paragraph{Consider the propagation of the concept names through the relational attributes.}

A specific naming challenge with RCA comes from the propagation of a concept across FC via the relational contexts, that leads to export the name of a concept from its native structure (e.g. a lattice) to related structure through relational attributes. 
When the relational attribute is considered as primary information, the propagated concept name may influence the name assigned to the concept under consideration. 
This propagation creates a chain of dependency in naming, which can result in an ambiguous, poorly chosen, or overly long name for a concept in a relational attribute leading to lexical inconsistency, increased verbosity, and structure maintenance issues.
In this sense, RCA turns concept naming into a global consistency problem, rather than a purely local labeling task.
E.g., in Fig. \ref{fig_pizzanora}, concept \texttt{CIng11} contains the attribute vegan, \texttt{CPizza31} refers to \texttt{CIng11} via the relational attribute based on  \texttt{contains}, and \texttt{CPizzer36} to \texttt{CPizza31} using the relation \texttt{serves}. The last two do not contain any other attributes. Let us suppose that \texttt{CIng11} is named \texttt{Legumes-only}, because vegan diet does not include grain. Considering this name for naming  \texttt{CPizzer36} could lead it to be named \texttt{Legumes-only serving}. This naming is inconsistent with the primitive attribute vegan, and thus creates a misunderstanding of the data.

\paragraph{Define an appropriate Name length.}

In both FCA and RCA, a concept name can be excessively long when it results from the aggregation of descriptive features. 
In FCA, a long descriptive name may correspond to a long intent made of several attributes, in which the label of each one may include its taxonomy or its category (e.g. for scaling). In RCA, this aspect is amplified by the scaling operators and the relations used to provide a relational attribute.
Such verbose labels may preserve information, but they reduce readability, especially in visualization frames exploration, and hinder downstream analytical tasks.
E.g. \texttt{CPizzer29} in Fig. \ref{fig_pizzarai} could be named \texttt{Dine-In White-Spring-Ingredient Red Vegetarian Summer-Ingredient Pizza Pizzeria}. But this naming, that does not include scaling operators and relation, mixes primary and secondary information as well as different constraints.
A more reasonable name could be \texttt{Vegetarian-Friendly Dine-In Pizzeria} as well as 
\texttt{indoor pizzeria vegetarian friendly}.
\\The underlying theoretical background for restricting name length and structure is the following. (i) A concept name is generally a noun phrase (and not a sentence).  (ii) The name has to avoid overloading the verbal memory of the designers and experts who will operate and maintain the resulting model. In this sense, Miller \cite{miller1956} first showed that the human memory buffer was limited to 7 plus or minus 2 items. The experimental setting was mostly relying on memorizing digits or simple words. Much later, Cowan \cite{Cowan2001}  restricted Miller's results to 4 elements. (iii) The cognitive load theory, as explained by Sweller \cite{Sweller1988}, points at the fact that a human processor, when dealing with problem solving, such as modeling analysis, must not be cognitively overloaded with a heavy linguistic processing, that is not central to the task. Limiting a name to the basic memory buffer will not unduly burden the task dedicated cognitive effort. So, psycholinguistic results tend to guide us toward a naming outcome between 3 and 5 linguistic units 
at most (either words or pairs of words tied with “-"). A single unit might be appropriate for a high-level concept (in the upper nodes of the graph). But, the more one goes down, the more a multiword noun phrase would be relevant (according to the hypernymy / hyponymy relation between concepts and their parents), restricted, as discussed before, to at most five units. 
Therefore, when naming a concept, we recommend limiting the length of the name to four units for English, and five for more articulated languages that have a greater use of prepositions (e.g. French). 


\section{A framework for handling variability in concept naming}
\label{sec_naming}

The \textbf{framework} is defined in the spirit of product lines \cite{DBLP:books/daglib/0015277}. It consists in determining a domain modeling step where the following elements are specified: (1) Variability sources and their constraints shaped as a model, prompt chunks that will be used to create a prompt corresponding to a selection of options (domain engineering in SPL vocabulary); 
(2) a prompt building engine using a user's options selection, respecting the above mentioned constraints (application engineering in SPL vocabulary). 

This section first 
introduces the language used to express variability (Sect. \ref{sec_uvl}), before describing the main sources of variability and potential strategies for naming FCA and RCA concepts (Sects. \ref{sec_sources} and \ref{sec_varstrategies}). It finally presents the operational pipeline integrating domain and application engineering in which  prompts are derived for suggesting concept names (Sect. \ref{sec_pipeline}).

\subsection{Modeling semantic variability in concept naming}
\label{sec_uvl}

Concept naming involves a set of semantic choices that are not fixed by the formal concept structure itself. A name may be based on the intent, the extent, inherited descriptors, neighboring concepts, implications, or relational attributes. In RCA, this variability is amplified because relational attributes in one  formal context may refer to concepts from other formal contexts, so that naming decisions may propagate across a family of concept structures.
Naming a concept therefore requires deciding which symbolic information should be used to determine it.

We model the sources of variability and the associated naming strategies using the Universal Variability Language (UVL\footnote{\url{https://universal-variability-language.github.io/}}) \cite{BENAVIDES2025112326}. 
In UVL, variability is expressed through feature models organized as logical tree structures, complemented with constraints between features. For the Boolean part used in this work, a UVL feature model can be interpreted in propositional logic: each feature corresponds to a Boolean variable, and feature relations and constraints restrict the combinations of variables that define valid configurations. This interpretation supports formal operations on configurations, such as satisfiability checking, configuration validation, and configuration counting.

From a notational point of view, UVL is a human-readable textual language designed by the Software Product Line (SPL) community to standardize variability modeling languages and to serve as a pivot format for variability-management tools. It also has a graphical representation based on feature models. 
In this graphical representation, nodes represent features and edges represent hierarchical refinement relations. Edge decorations specify variability constraints: a direct line ending in a small solid or hollow circle indicates, respectively, that a feature is mandatory or optional; a set of lines drawn within a semicircle connected to the source feature groups features according to an ``OR'' or ``XOR'' relation, the latter being called an alternative in UVL. Additional propositional constraints can further restrict the set of valid configurations, for example by expressing mutual exclusions.

The variability models of this section can be read and manipulated within tools compatible with UVL, e.g., the web application flamapy\footnote{\url{https://ide.flamapy.org/editor}} which also provides classical operations on models, including the formal operations mentioned above, and support for practical operations such as interactive configuration building.  
Such tools can be integrated into the operational part of the proposed framework to check, derive, and compare configurations before they are used to generate naming prompts.

These configurations are then used in our framework to generate prompts for an LLM, which suggests candidate names. When applied to real-world data, the suggested names should be validated by an expert, which may involve human--LLM interaction. 

\subsection{Sources of variability associated with the concepts}
\label{sec_sources}

The variability in naming a concept can be provided using its content, the concept graph structure (e.g. lattice), and the related concept graph structures  when using RCA. The set of variability sources and strategies presented in this section is not intended to be exhaustive and is based primarily on solutions drawn from the literature \cite{arandaConcepts2024,DBLP:conf/concepts/GuenouneGHLMMZ25}. 
It can easily be extended thanks to the approach based on SPL principles.

\begin{figure}[htb]
    \centering
\includegraphics[width=\textwidth]{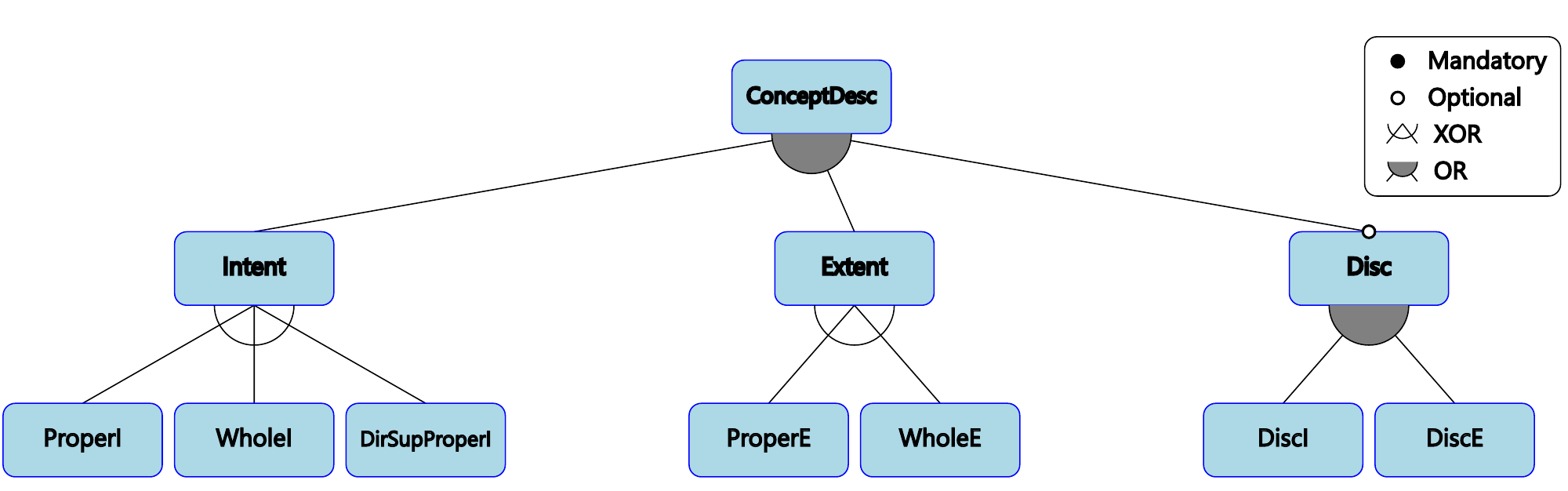}
\caption{Variability in the description of a concept using UVL (figure built with flamapy).}
\label{fig_ConceptDescrFM}
\end{figure}

\paragraph{Variability from concept description.}
Various information can be used to describe a concept, i.e. its intent, its extent, both, or discriminating attributes or objects to be favored (see Fig. \ref{fig_ConceptDescrFM}). Regarding the intent, the information to provide can be introduced attributes (\texttt{ProperI}), introduced and inherited attributes (\texttt{WholeI}), or attributes introduced in the direct super-concepts (\texttt{DirSupProperI}). For the extent, information can be the introduced objects (\texttt{ProperE}) or the introduced and inherited objects (\texttt{WholeE}). The discriminating attributes or objects can be provided by the concept itself (\texttt{DiscI}) or the direct super-concepts (\texttt{DiscE}).
\\For concept naming, using the intent, the extent, or both  defines the lexical domain from which the name will be provided. Using only the intent, the naming is pushed toward definition-based, stable names that reflect shared properties and relational constraints of the concept. The concept name then remains robust when the extent changes. 
Using only the extent, one tends to produce exemplar or instance driven names (e.g., using salient object labels), which can be intuitive but often brittle, dataset-specific, and sometimes misleading when objects are atypical. 
Using both intent and extent usually yields the most human-friendly names because the intent is used for the formal meaning and the extent as concrete ``anchors''. This approach increases the risk of overemphasizing important instances and disclosing object names into the concept label.
A good compromise may be to use both while explicitly constraining the naming rule in the strategy.
\\A key choice is whether to use proper or whole information. Using \texttt{ProperI}  highlights what is specific to the concept and often results in more precise and discriminating names. But this can result in the loss of important contextual information that is transmitted. Using \texttt{WholeI} is more faithful to the full meaning of the concept in the lattice and produces more stable, self-contained descriptions, at the cost of longer and sometimes less distinctive names. The same trade-off holds for the extent: \texttt{ProperE} focuses on the objects uniquely captured at that node (more contrastive), while \texttt{WholeE} reflects the full set of objects covered by the concept (more representative). 
In \cite{DBLP:conf/concepts/GuenouneGHLMMZ25}, \texttt{WholeI} and \texttt{WholeE} are used, while distinguishing between proper and inherited in the data given to the LLM. Other variants are using attributes introduced in direct super-concepts (\texttt{DirSupProperI}) as in \cite{aranda_corral_2026_18608706}.
\\The use of discriminating attributes or objects further reduces the lexical domain, e.g. using the objects in the extents of the concept’s direct super-concepts (\texttt{DiscE}) but not in the concept itself as in \cite{aranda_corral_2026_18608706},  and similarly using the attributes taken from the intents of its direct subconcepts and its sibling concepts but not in the concept itself (\texttt{DiscI}).

\paragraph{Variability from the structure of concepts.}

\begin{figure}[htb]
    \centering
\includegraphics[width=0.5\textwidth]{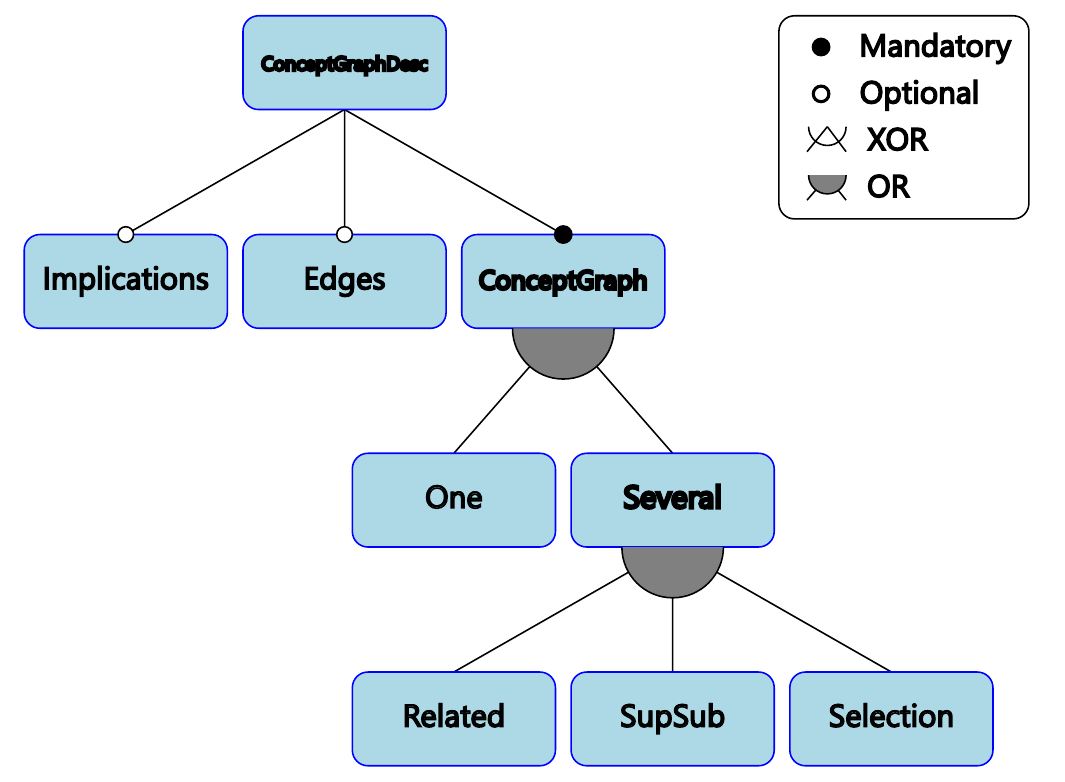}
\caption{Variability in the description of a graph of concepts described using UVL (figure built with flamapy). Two propositional-logic constraints are  applied to this tree structure: \texttt{Edges $\Rightarrow$ Several} and \texttt{SupSub $\Rightarrow$ Edges}.}
\label{fig_ConceptGraphDesc}
\end{figure}

Using exclusively the concept to determine its name (as adopted in \cite{DBLP:conf/concepts/GuenouneGHLMMZ25}) provides a reduced lexical domain made solely of the concept's own description (intent and/or extent). This is lightweight and often sufficient, but it can yield generic labels and offers limited support for ensuring consistency across a family of concepts. Alternatively, providing several concepts enlarges the lexical domain using contrasted context, and enables naming all the concepts jointly and coherently. This matters in FCA but is specifically relevant for RCA: when a concept intent contains relational attributes pointing to other concepts, then including the referenced concepts in the entry allows the proposed names to be reused to create compositional labels (e.g., “Pizzerias serving [Pizza-concept name]”, “Pizzas containing only [Ingredient-concept name]”). Naming a set together therefore could improve both semantic accuracy and terminological consistency, at the possible cost of greater semantic dispersion from the LLM’s perspective.
\\Three sources of variability can be provided in the concept graph description, i.e., the implications related to the concepts to be named (\texttt{Implications}), edges related to the concepts (\texttt{Edges}), and the graph of concepts (\texttt{ConceptGraph}) as presented in Fig. \ref{fig_ConceptGraphDesc}. 
For the latter, \texttt{One} and \texttt{Several} correspond respectively to the concept to be named and several concepts to be named together. They can be the related concepts described in other graphs  when using RCA (\texttt{Related}), the direct super and  sub-concepts (\texttt{SupSub}), and selected ones according to criteria that the analyst has to specify  (\texttt{Selection}).
Using \texttt{SupSub} helps to identify a right abstraction level for the name. \texttt{Related} is suited to RCA: it includes the key concepts targeted by the most important relational attributes, so that relational semantics can be expressed directly in the name,  and names can be propagated. Finally,  \texttt{Selection} supports task-driven naming by restricting context to the concepts relevant for a given report, query, or user focus. \texttt{Edges} and \texttt{Implications} are optional enhancements that allow for further refinement of the concept naming. Using  \texttt{Edges} makes the relations between the provided concepts explicit; this is especially helpful when multiple concepts are named together and improves consistency. Using \texttt{Implications} tends to shift attention toward the most informative attributes by highlighting what is entailed; this can improve stability and reduce reliance on objects and incidental features.

\begin{figure}[htb]
    \centering
    \includegraphics[width=0.35\textwidth]{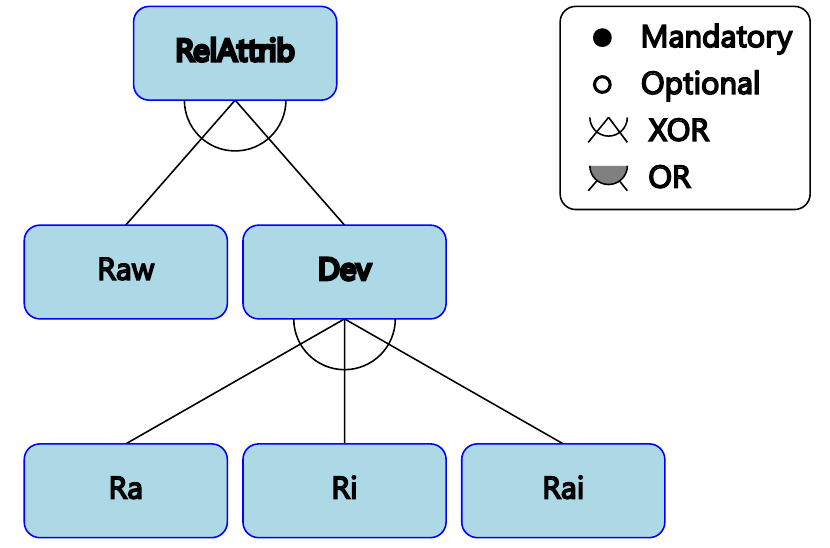}
\caption{Variability in the rewriting of relational attributes described using UVL (figure built with flamapy).}
\label{fig_raGraph}
\end{figure}

\paragraph{Variability from  relational attribute.}
The name of a relational attribute provided in a concept can be more or less explicit depending on whether it includes the identifier (e.g. C\_Pizza\_21) of the related concept (\texttt{Raw}) or the attributes it conveys (\texttt{Dev}, for ``Development''). Fig. \ref{fig_raGraph} presents the sources of variability for relational attribute rewriting. In the case of \texttt{Raw}, relevant connected concepts should be given as input. 
Regarding \texttt{Dev}, three variants are proposed in \cite{DBLP:conf/concepts/GutierrezHMZ25} to rewrite the target concept. \texttt{Ra} (standing for ``rewriting with proper attributes'') provides only its introduced/proper attributes. When  
the concept has no introduced attributes, it is left as a concept identifier.
For the naming, \texttt{Ra} keeps relational cues concise and avoids injecting overly generic inherited information. It should be preferred when short, discriminant names are expected.
\texttt{Rai}, standing for ``rewriting using proper attributes, alternatively use inherited'', provides the introduced attributes, and the inherited ones when there are none (cf. Section \ref{sec_rca} for an illustration). Using the latter provides a fallback description that prevents ``empty'' targets, thereby improving interpretability while promoting the specific clues introduced.
Finally, \texttt{Ri}, which stands for ``rewriting with inherited'', includes the overall intent in relational attributes, whether they are introduced or inherited attributes. For naming, this maximizes the lexical domain and can help when full meaning matters, but tends to be verbose and may dilute the most discriminating terms.

\subsection{Variability in strategies for naming a concept}
\label{sec_varstrategies}

\begin{figure}[htb]
    \centering
\includegraphics[width=\textwidth]{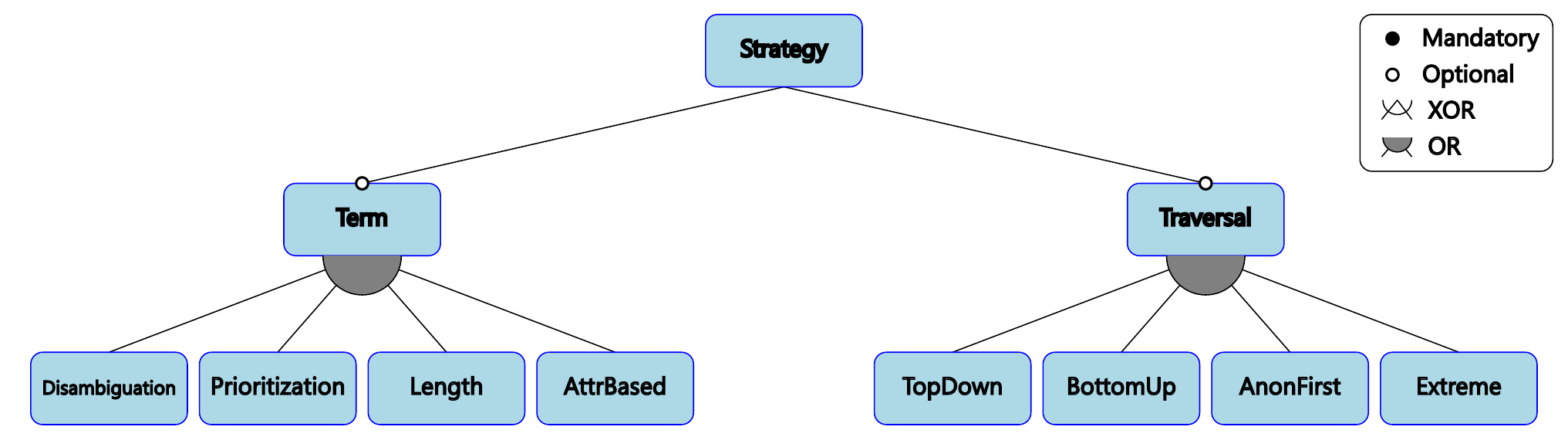}
\caption{Variability in the description of a concept naming strategy using UVL (figure built with flamapy).}
\label{fig_strategy}
\end{figure}

Two aspects can be taken into account when defining a concept naming strategy (cf. Fig. \ref{fig_strategy}).
The first aspect, named \texttt{Term}, concerns the linguistic unit to build. It is defined using four lightweight sources of variability.
First, word-sense discrimination (\texttt{Disambiguation}) handles each of the ambiguous labels of attributes, objects, and relations considered in the naming by inferring the intended meaning from the domain context, the descriptor type, and co-occurring descriptors; when ambiguity persists, a reasonable assumption must be made.
The second source of variability (\texttt{Prioritization}) distinguishes between primary information (salient, discriminating, stable) and secondary information, in order to build names from the primary descriptors. The use of secondary information makes it possible to resolve ambiguities or improve specificity. 
The third source of variability  (\texttt{Length}) constrains the output to a single candidate respecting a fixed template, e.g., “Adj Name1 Adj Name2", with a one-sentence rationale, and encourages shorter names for higher, more general concepts in the lattice. The fourth source of variability (\texttt{AttrBased}) considers  intent–extent guidance to enforce intent-driven naming: the intent is the primary signal, while the extent is used only for disambiguation and not as a source of object names.

The second aspect of variability in the strategy concerns the graph traversal. There are four ordering strategies for naming concepts, each offering a different advantage in terms of consistency and control of effort.
Naming concepts using \texttt{TopDown} traversal (following a topological sorting of the lattice) establishes a common vocabulary very early on. The aim is that general concepts get short, generic names first, and more specific concepts can then reuse or refine this terminology, improving global consistency.
Naming using \texttt{BottomUp} traversal starts with highly specific concepts whose intent/extent is rich, which often makes them easier to label. These concrete names can then be generalized when moving upward, but care is needed to avoid overly instance-driven naming.
\\Starting with concepts that have no introduced attributes (\texttt{AnonFirst}), i.e., the “anonymous” concepts in \cite{aranda_corral_2026_18608706}, helps fill gaps in the description as these concepts lack a direct linguistic handle. Assigning them names provides such handles and makes subsequent naming steps easier and more coherent. The authors of \cite{aranda_corral_2026_18608706} use these names almost like newly introduced attributes.
Finally, for \texttt{Extreme} cases, e.g., when the top concept  has an almost empty intent, instead of using a neutral placeholder (e.g. “Top" or “Thing"), grounding the name in the domain label (e.g., Pizzeria, Pizza, Ingredient) and refining it with the most specific generalization supported by the closest informative sub-concepts (typically the nearest nodes with introduced attributes) is a key to reduce variability. This yields a short but meaningful umbrella name.

\subsection{Operationalizing the variability model}
\label{sec_pipeline}

\begin{figure}[htb]
    \centering
\includegraphics[width=\textwidth]{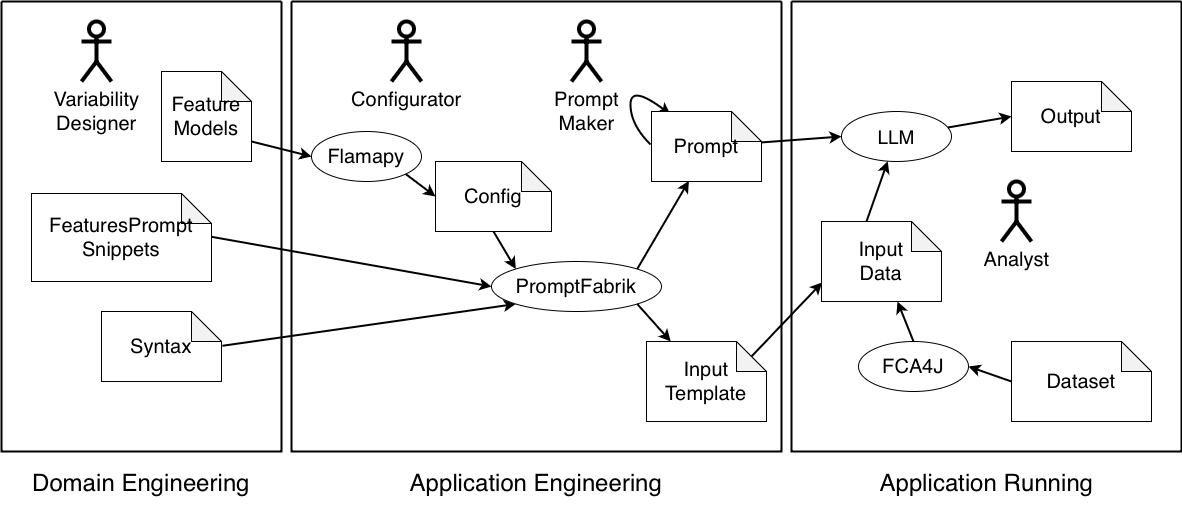}
\caption{Product Line pipeline. Oval and rectangular shapes represent respectively a software and a data file. }\label{fig_pipeline}
\end{figure}

The variability model defines a space of possible naming configurations. To make this space operational, it is integrated in a pipeline that derives LLM prompts from selected configurations. 
The pipeline, shown in Fig. \ref{fig_pipeline}, follows SPL principles. Variability and assets (prompt snippets for features and syntax) are defined at the domain engineering step. At the application engineering step, a prompt fabric assembles relevant prompt snippets,  according to a selected configuration,  to produce a prompt that will be run on selected input data (the prompt is here the application).
The  pipeline enables a systematic exploration of how different sources of information from the concept structure influence concept naming. %
This subsection first introduces the structure of a naming prompt (target of the pipeline), then describes the pipeline, before presenting its main elements.

\paragraph{Structure of a concept naming prompt.}

\begin{figure}
    \centering
    \includegraphics[width=0.5\textwidth]{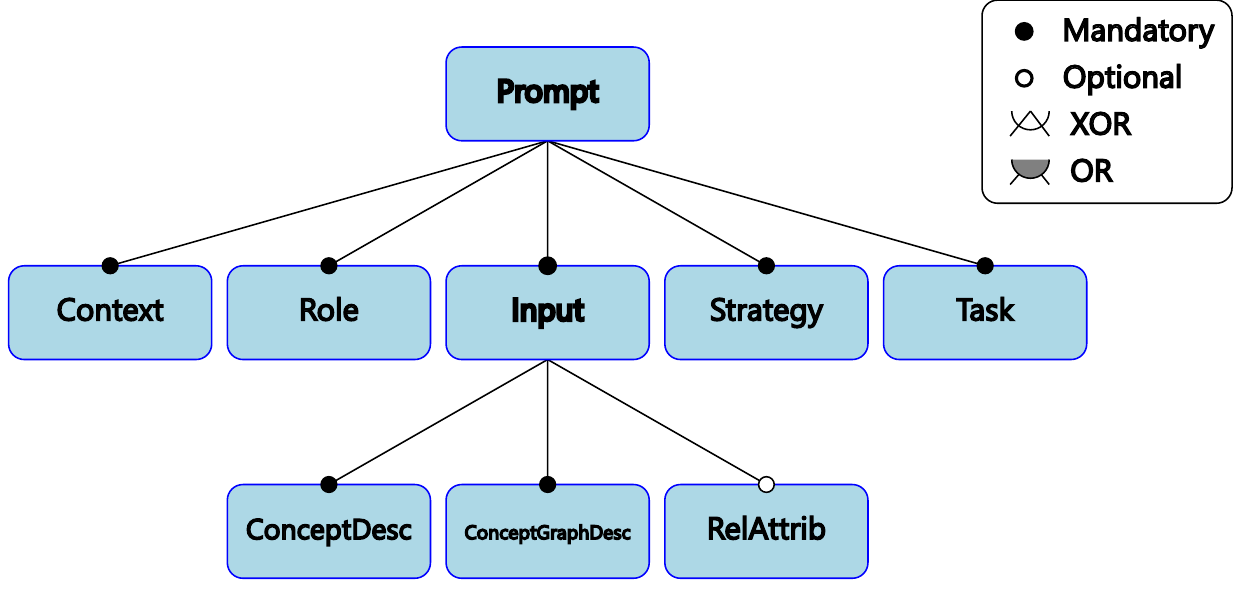}
    \caption{Prompt variability feature model 
    described using UVL (figure built with flamapy).}
    \label{fig_fmprompt}
\end{figure}

\begin{figure}[htb]
    \centering
\includegraphics[width=0.5\textwidth]{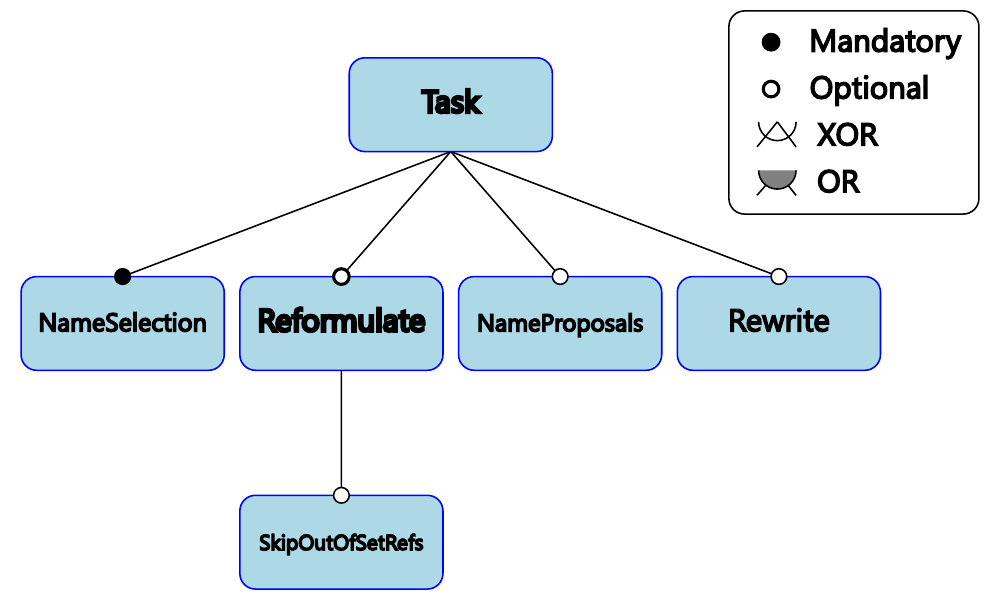}
\caption{Variability on tasks using UVL (figure built with flamapy).}
\label{fig_tasks}
\end{figure}

The structure of a naming prompt is made of five consecutive parts, i.e. the context of the task to be performed by the LLM (\texttt{Context}), the role of the LLM (\texttt{Role}), the input that is provided to the LLM to perform the task (\texttt{Input}), the expected strategy (\texttt{Strategy}), and finally the task to be performed (\texttt{Task}). Each of these parts may include  variability (cf. Fig. \ref{fig_fmprompt}). The variability for  
(\texttt{Context}) and (\texttt{Role}) is not explored in depth in this paper, as that is not its focus. The variability for  (\texttt{Input}) is described in the previous subsections, through Fig. \ref{fig_ConceptDescrFM}, \ref{fig_ConceptGraphDesc},  \ref{fig_raGraph} and \ref{fig_strategy}. The \texttt{Task} variability is presented in Fig. \ref{fig_tasks}. 
In the latter, \texttt{NameSelection} is the only mandatory task: for each concept, one final name must be chosen, typically among candidates and/or after expert discussion. The other tasks are optional and serve as support steps. \texttt{NameProposals} generates several candidate names per concept before selection. \texttt{Reformulate} provides a plain-language explanation of the concept to help justify and validate naming decisions; when reformulating a set of related concepts, \texttt{SkipOutOfSetRefs} may be activated to ignore expressions that refer to concepts not included in the provided set. Finally, \texttt{Rewrite} applies the chosen names back to the original representation by replacing concept identifiers and exporting the updated concept descriptions.

\paragraph{The prompt fabric pipeline.}

A Software Product Line (SPL) is typically structured into three complementary levels \cite{DBLP:books/daglib/0015277}: domain engineering, application engineering, and the running application. 
Domain engineering  builds the foundation of the product line by using domain commonalities and variation points to produce reusable artifacts such as feature models, reference architectures (e.g. configurable prompts), reusable components such as snippets, and generation assets (e.g., prompt fabric).
Application engineering then establishes a specific application from this common core by selecting a coherent set of features, configuring reusable assets, and developing application-specific artifacts, such as assembled components, parameter settings, and input data.
Finally, the application running level corresponds to the deployed and executable instance of the established application: it is the operational realization of a particular product.
There are two main approaches for implementing SPL, i.e.,  the annotative one, which represents the variability by marking a shared artifact with presence conditions or annotations, and the compositional approach. The latter realizes variability by assembling distinct artifacts or modules according to a selected configuration. The implemented SPL approach is defined in the domain engineering level, where variability is modeled. Figure \ref{fig_pipeline} presents the prompt fabric pipeline designed for this paper according to SPL principles. 
It is mainly based on an annotative approach for the core aspects, while relying partly on a compositional approach for input syntax.

\paragraph{The elements of the pipeline}
The domain engineering comprises three text files. The file \texttt{FeatureModels} lists all variability features in a tree structure in UVL syntax. 
At the end of the file, a “constraints” section enables listing propositional-logic constraints applied to this tree structure.
The file \texttt{Features\-Prompt\-Snippets} is the core  annotated prompt template containing all the prompt snippets, except those relative to syntax (e.g. dot format file\footnote{Format for Graphviz tool \url{https://graphviz.org/}.}). 
A prompt snippet
is referenced by the name of a feature from the feature model (cf. Fig. \ref{fig_ConceptDescrFM}, \ref{fig_ConceptGraphDesc},  \ref{fig_raGraph}, \ref{fig_strategy}, \ref{fig_fmprompt}, and \ref{fig_tasks}) and it may refer to external data, such as an input file or the snippets describing syntax. As presented in the file excerpt presented below, a feature snippet reference is declared after the characters \#\#  and external information are written in italics.

\begin{promptbox}[Excerpt from file FeaturesPromptSnippets]
\footnotesize{
\#Prompt snippets associated to the different Features

\#\#Feature Input

Inputs that will be taken from the *input file* (`Input.md`) appear in **input[[X]]**, where **X** is variable. 
Then they are referenced through **X**.

\#\#Feature Context

You help name formal concepts in Formal Concept Analysis (FCA).

The domain is **input[[Domain description]]**.

\#\#Feature Role

You are a terminology-aware assistant.
You produce short, domain-appropriate names that are consistent across similar concepts.

\#\#Feature Input/ConceptDesc

\textit{**syntax[[Feature Input/ConceptDesc]]**.}

(...)

\#\#\#Feature Input/ConceptDesc/Intent/WholeI

The concept description contains all the concept attributes (introduced and inherited).

\textit{**syntax[[Feature Input/ConceptDesc/Intent/WholeI]]**}

(...)

\#\#Feature Input/ConceptGraphDesc/ConceptGraph/One

For your analysis, here is the concept description to be used: 
\textit{**input[[Concept description]]**.}

\#\#Feature Input/ConceptGraphDesc/ConceptGraph/Several

For your analysis, here is the concept set description to be used: \textit{**input[[Concept set description]]**.}

You should assign a name to all concepts of the set.

\#\#Feature Input/ConceptGraphDesc/ConceptGraph/Several/Related

The given concepts are connected through relational attributes. This means that in the intent of concept C1, you can find a relational attribute q\_r(C2) where C2 is another concept.\\
(...)
}
\end{promptbox}

The file \texttt{Syntax} contains prompt snippets for features using specific syntax as inputs, such as the description of a file format. 
For instance, the excerpt below explains the syntax to the LLM to read an input file, in dot format, for feature \texttt{ConceptDesc} (cf. Fig. \ref{fig_ConceptDescrFM}). 

\begin{promptbox}[Excerpt from file Syntax]
\footnotesize{\#Prompt snippets 

\#\#Feature Input/ConceptDesc

The concept description comes from a file created for Graphviz, in .dot format. It corresponds to a node of a graph, which represents a concept. The node is described by a text between characters ‘[‘ and ‘];'. This text contains a label between characters ‘’ and ‘’. 
The label is composed of three parts: identification, intent, extent. \\
-	The first part is an identifier, and is finished by character ‘|’. \\
-	The next part (intent), also finished by character ‘|’, is a list of attributes separated by special character \verb|\n|. \\
-	The next part (extent) is a list of objects, also finished by character ‘|’. In the Domain, they correspond to objects sharing the attributes of the intent.\\
(...)\\
\#\#\#Feature Input/ConceptDesc/Intent/WholeI\\
(...)\\
}
\end{promptbox}

In the application engineering level, the use of an UVL application such as flamapy, enables creating the variability configuration file \texttt{Config} that lists selected features from \texttt{FeatureModels}. The benefit of using such an application is that it guides the configuration process based on the constraints defined in the feature model. An excerpt of \texttt{Config} is presented below.  

\begin{promptbox}[Excerpt from file Config]
\footnotesize{
\# Configuration profile\\
\#\# Feature selection\\
- Input\\
- Context\\
- Role\\
- Input/ConceptDesc\\
- Input/ConceptDesc/Intent/WholeI\\
- Input/ConceptDesc/Extent/WholeE\\
- Input/ConceptGraphDesc/ConceptGraph/Several\\
(...)
}
\end{promptbox}

Using \texttt{Config} as entry, the software program
\texttt{PromptFabrik}, developed for this SPL, then weaves the two snippet files, i.e. \texttt{FeaturesPromptSnippets} as core and \texttt{Syntax} that contains syntax description, to generate the prompt file \texttt{Prompt}. This composition consists of injecting the syntax into the core file keeping exclusively snippets listed in \texttt{Config}. In the generated prompt excerpt presented below, as feature \texttt{One} (cf. Fig. \ref{fig_ConceptGraphDesc}) has not been selected, the associated snippet does not appear. To make it easier to read, the inserted syntax parts are written in italics. At this stage, \texttt{Prompt} may be refined by a prompt maker (usually a developer). In this SPL, no  refinements are expected.

\begin{promptbox}[Excerpt from generated file Prompt]
\footnotesize{
\# Prompt

\#\# Context

You help name formal concepts in Formal Concept Analysis (FCA).

The domain is **input[[Domain description]]**.

\#\# Role

You are a terminology-aware assistant.
You produce short, domain-appropriate names that are consistent across similar concepts.

\#\# Input

Inputs that will be taken from the *input file* (`Input.md`) appear in **input[[X]]**, where **X** is variable. 
Then they are referenced through **X**.

\#\# Input/ConceptDesc

\textit{The concept description comes from a file created for Graphviz, in .dot format. It corresponds to a node of a graph, which represents a concept. The node is described by a text between characters ‘[‘ and ‘];'. This text contains a label between characters ‘’ and ‘’.}

\textit{The label is composed of three parts: identification, intent, extent.}

\textit{-	The first part is an identifier, and is finished by character ‘|’.}

\textit{-	The next part (intent), also finished by character ‘|’, is a list of attributes separated by special character} \textit{\textbackslash n}.

\textit{-	The next part (extent) is a list of objects, also finished by character ‘|’. In the Domain, they correspond to objects sharing the attributes of the intent.}

\#\# Input/ConceptDesc/Intent/WholeI

The concept description contains all the concept attributes (introduced and inherited).

\textit{The special term ‘/\_INH\_ATT\_/’ marks the boundary in the intent: the introduced attributes (proper to objects of the concept) appear before, and the attributes inherited from higher concepts in the lattice appear after it.}

\#\#  Input/ConceptGraphDesc/ConceptGraph/Several

For your analysis, here is the concept set description to be used: **input[[Concept set description]]**.

You should assign a name to all concepts of the set.

(...)
}
\end{promptbox}

\texttt{PromptFabrik} also generates the input data file template \texttt{InputTemplate}. This template is based on the selected variability features of the input (Fig. \ref{fig_ConceptDescrFM}, \ref{fig_ConceptGraphDesc},  and \ref{fig_raGraph}),  
and thus informs on the expected input data required for concept naming. An excerpt of this file is provided below. 

\begin{promptbox}[Excerpt from generated file InputTemplate]
\footnotesize{
\# InputTemplate

\#\# Domain description

[Write the value for Domain description here.]

\#\# Concept set description

[Write the value for Concept set description here.]

}
\end{promptbox}

In the Running application level, the file \texttt{InputData} must be created using the template file \texttt{InputTemplate} and the data describing concepts to be named. The data are generated using FCA4J~\cite{DBLP:conf/cla/GutierrezH022}. The excerpt from \texttt{InputData}, presented below, lists concepts built by RCA and presented in Fig. \ref{fig_pizzanora}. Concepts are described in dot/Graphviz syntax. Finally, the application is run by providing the files \texttt{InputData} and \texttt{Prompt} to the LLM, which suggests concept names. Depending on the result, the analyst (e.g. domain expert) may then interact with the LLM to refine the concept name.

\begin{promptbox}[Excerpt from file InputData]
\footnotesize{
\# InputTemplate

\#\# Domain description

The domain describes pizzerias that serve pizzas that contain ingredients.

\#\# Concept set description

```
Ingredient concepts

9 [shape=record,style=filled,fillcolor=lightblue,label="\{C\_Ingredient\_9 \textbar \textbar  \_INH\_OBJ\_\verb|\n| goatcheese\verb|\n| burrata\verb|\n| scallop\verb|\n| tomato\verb|\n|shallot\verb|\n| mushroom\verb|\n| eggplant\verb|\n|\}"];\\
(...)\\
Pizza Concepts\\
18 [shape=record,style=filled,fillcolor=lightblue,label="{C\_Pizza\_18 \textbar\\ existForall\_contains(\_ALL\_OBJECTS\_)\verb|\n| \textbar\_INH\_OBJ\_\verb|\n| chevrette\verb|\n| forest\verb|\n| violet\verb|\n| stjacques\verb|\n|}"];\\
(...)\\
Pizzeria Concepts\\
24 [shape=record,style=filled,fillcolor=lightblue,label="\{C\_Pizzeria\_24 \textbar \\ exist\_serves(existForall\_contains(\_ALL\_OBJECTS\_))  \verb|\n| exist\_serves(existForall\_contains(vege))\verb|\n| \\ exist\_serves(existForall\_contains(summer))\verb|\n| \textbar\_INH\_OBJ\_\verb|\n| happizzy\verb|\n| eataly\verb|\n| lafelicita\verb|\n| smallitaly\verb|\n|\}"];\\
(...)
}
\end{promptbox}


\section{Illustration with the Pizzeria model}
\label{sec_evaluation}

This section illustrates how the proposed framework can be instantiated to explore naming variability on the Pizzeria relational dataset introduced in Section \ref{sec_basics}. 
The generated concept lattices are those shown in Figure \ref{fig_pizzanora} and result from the relational context family described in Table~\ref{table_pizzas}.

Since several LLMs exist, and each run is quite costly in time, we decided to restrict our test to those LLMs most able to provide correct answers. In a previous work \cite{DBLP:conf/concepts/GuenouneGHLMMZ25}, at least five LLMs have been tested, in an attempt to perform a first naming on much simpler models with less than 15 concepts. Two LLMs outperformed the others and offered interesting tracks, Claude Sonnet 4.6 and Chatgpt 5.2 Thinking. When moving to the Pizzeria model which is up to 36 concepts, only the best LLMs were kept, since  several configurations were to be run, which is quite time consuming.
This proof-of-concept illustration makes it possible to observe whether the proposed configurations lead to similar or different naming behaviors across systems. The goal is not to benchmark LLMs, but to illustrate how the variability framework can control the information provided to a naming assistant.

This section first presents the selected configurations, then discusses the resulting naming variations, and finally reports a small expert-based assessment of selected names.

\subsection{Selected configurations}
\label{sec_config}

The selected configurations instantiate distinct points in the variability space. They are chosen to illustrate how neighboring concepts and relational attributes may influence the names suggested for a concept.
Thus, two comparative sets of runs have been made: 
 one where a concept is named independently of the others, using its own information,
and one where all the concepts are named at the same time.
We focus on the assessment of the concepts \texttt{CPizzer35} and \texttt{CPizzer36} of the pizzeria lattice as each one has no primitive attributes, introduces only a relational attribute, has children and siblings, and is not at the top of the lattice.

Three combinations of features, each combination corresponding to a configuration, were tested. The three configurations present similar features to describe a concept  (\texttt{ConceptDesc})  
in the input of the prompt: using the whole intent (\texttt{WholeI}) and extent (\texttt{WholeE}).
In terms of strategy, all configurations use the four sources of variability for \texttt{Term}, namely \texttt{Disambiguation}, \texttt{Prioritization}, \texttt{Length}, and \texttt{AttrBased}. 
They also have to perform the same tasks (cf. Fig. \ref{fig_tasks}), i.e. to suggest several candidate names (\texttt{NameProposals}) and select one, indicating the choice rationale (\texttt{NameSelection}). The configurations differ exclusively in some variability features of the prompt input. For the configuration that suggests a name for one concept (independently of others): (i) no information on the graph is provided, corresponding to the variability feature \texttt{One} 
and (ii) the relational attributes are developed using \texttt{Rai}.  
For the two other configurations, the variability feature to describe a graph \texttt{ConceptGraphDesc} is \texttt{Related}. They differ in the development of relational attributes, i.e. one uses \texttt{Rai} and the other \texttt{Raw}, corresponding respectively to its complete rewriting and no rewriting. 
The output of the two LLMs, i.e. ChatGPT 5.2 Thinking and Claude Sonnet 4.6 Extended web applications (both used on March 6, 2026), are used to observe how the same configurations may lead to convergent or divergent naming suggestions.
For each configuration, two runs were conducted to assess LLM stability in the name-suggestion process, for a total of 16 trials across all configurations\footnote{Data are available at \url{https://seafile.lirmm.fr/d/884f1a48642f4c7a909b/} password: ConceptNaming, and as ancillary files of the Arxiv paper.}.

\subsection{Qualitative Analysis}

This subsection first analyzes the results for two concepts (\texttt{CPizzer35} and \texttt{CPizzer36}) in all configurations (\texttt{One},  \texttt{Several/Rai} and \texttt{Several/Raw}).

\begin{figure}[htb]
\centering
  \includegraphics[width=\textwidth]{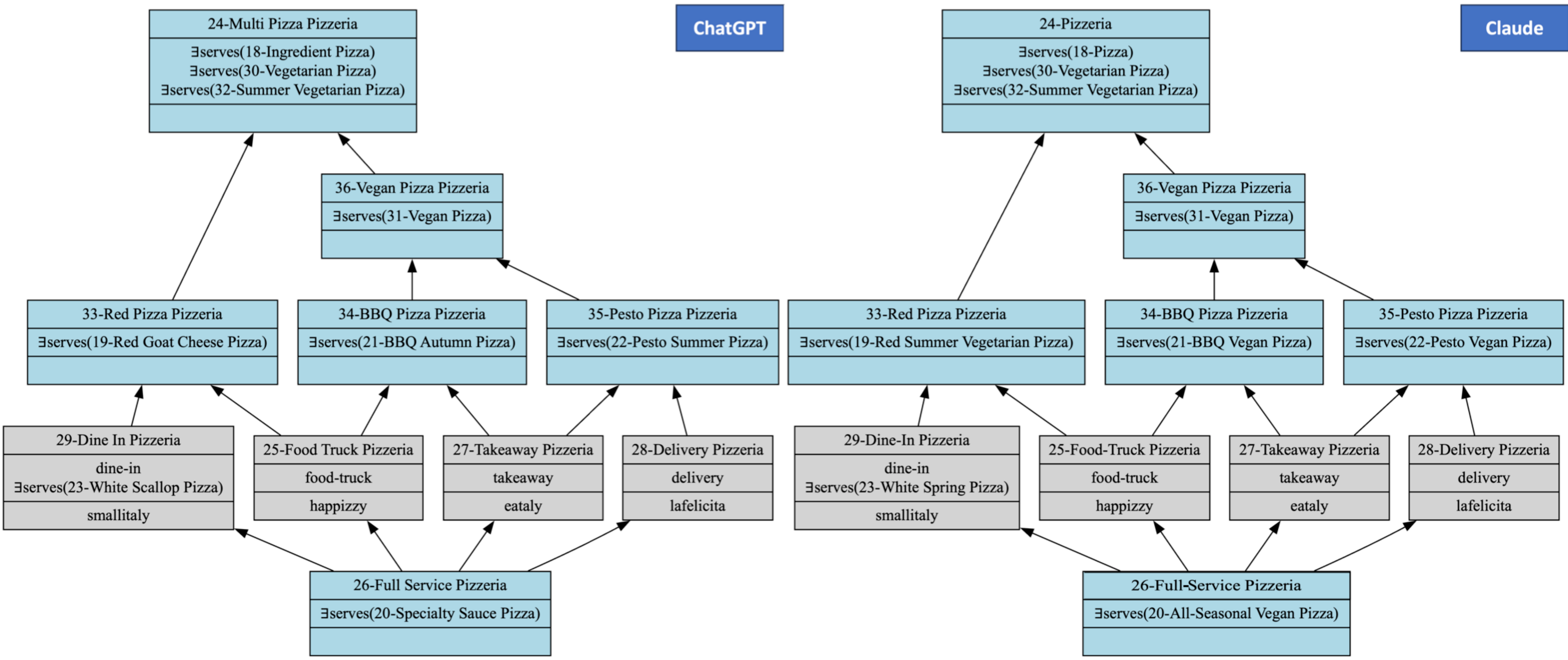}
  \caption{Pizzeria lattices from Trials 14 (left-hand side) and 15 (right-hand side). The number preceding the concept name corresponds to the concept ID in Fig \ref{fig_pizzanora}.
  }
  \label{fig_PizzeriasChatGPTClaude}
\end{figure}

\begin{figure}[htb]
\centering
\includegraphics[width=0.75\textwidth]{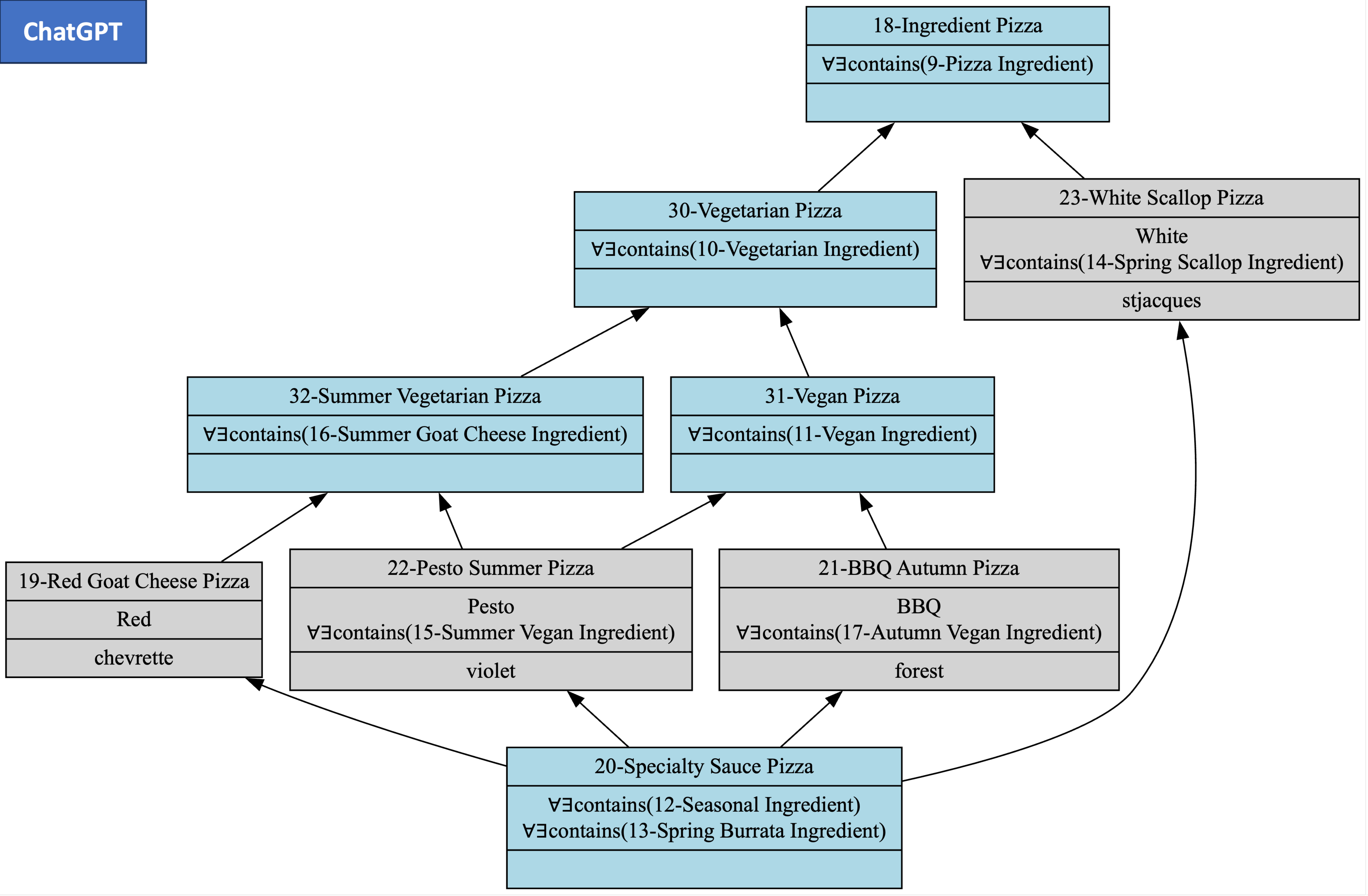}
\includegraphics[width=0.75\textwidth]{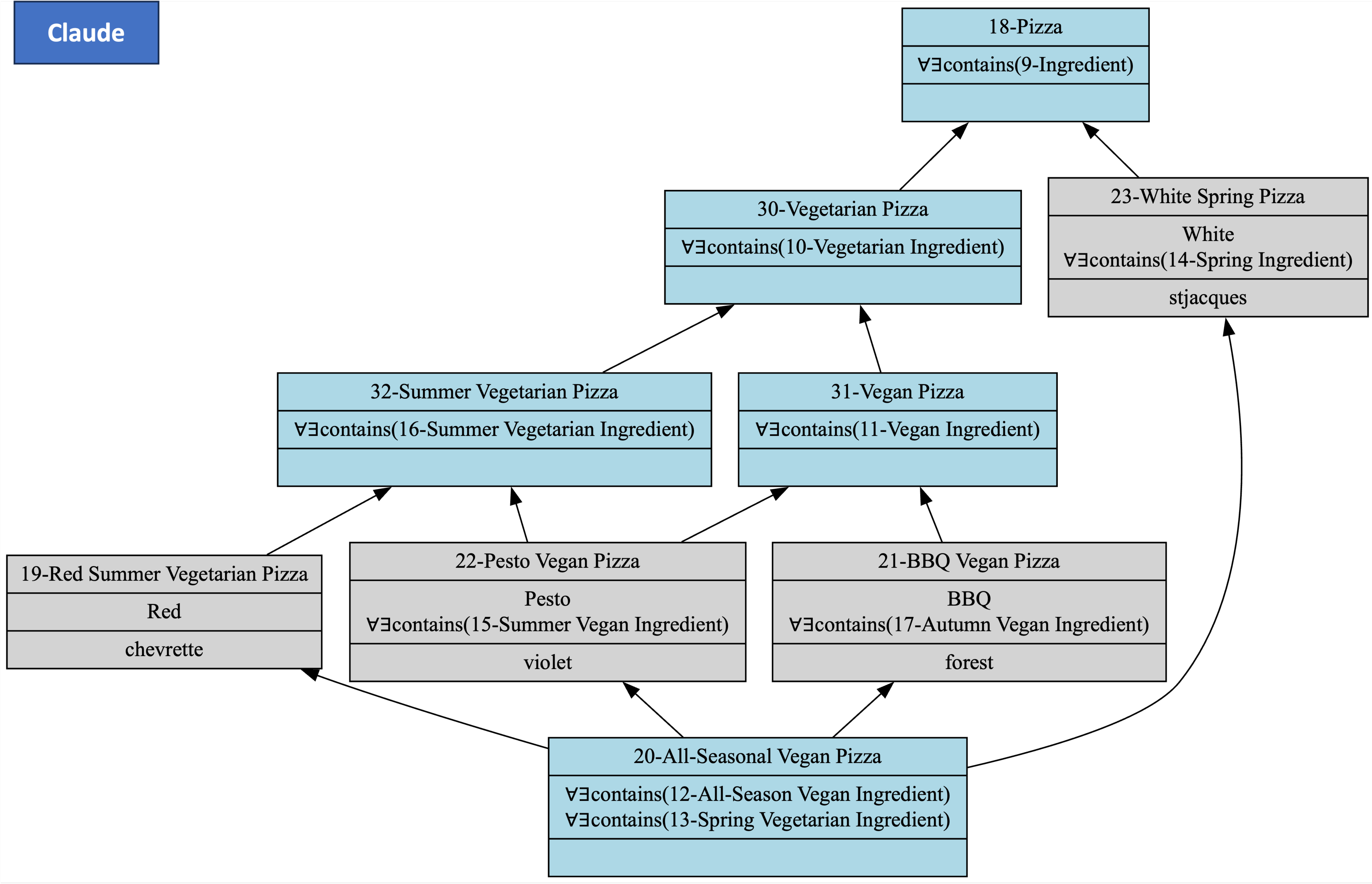}
  \caption{Pizza lattices from Trials 14 (Top) and 15 (Bottom). The number preceding the concept name corresponds to the concept ID in Fig \ref{fig_pizzanora}.
  }
  \label{fig_PizzasChatGPTClaude}
\end{figure}

\begin{figure}[htb]
\centering
  \includegraphics[width=\textwidth]{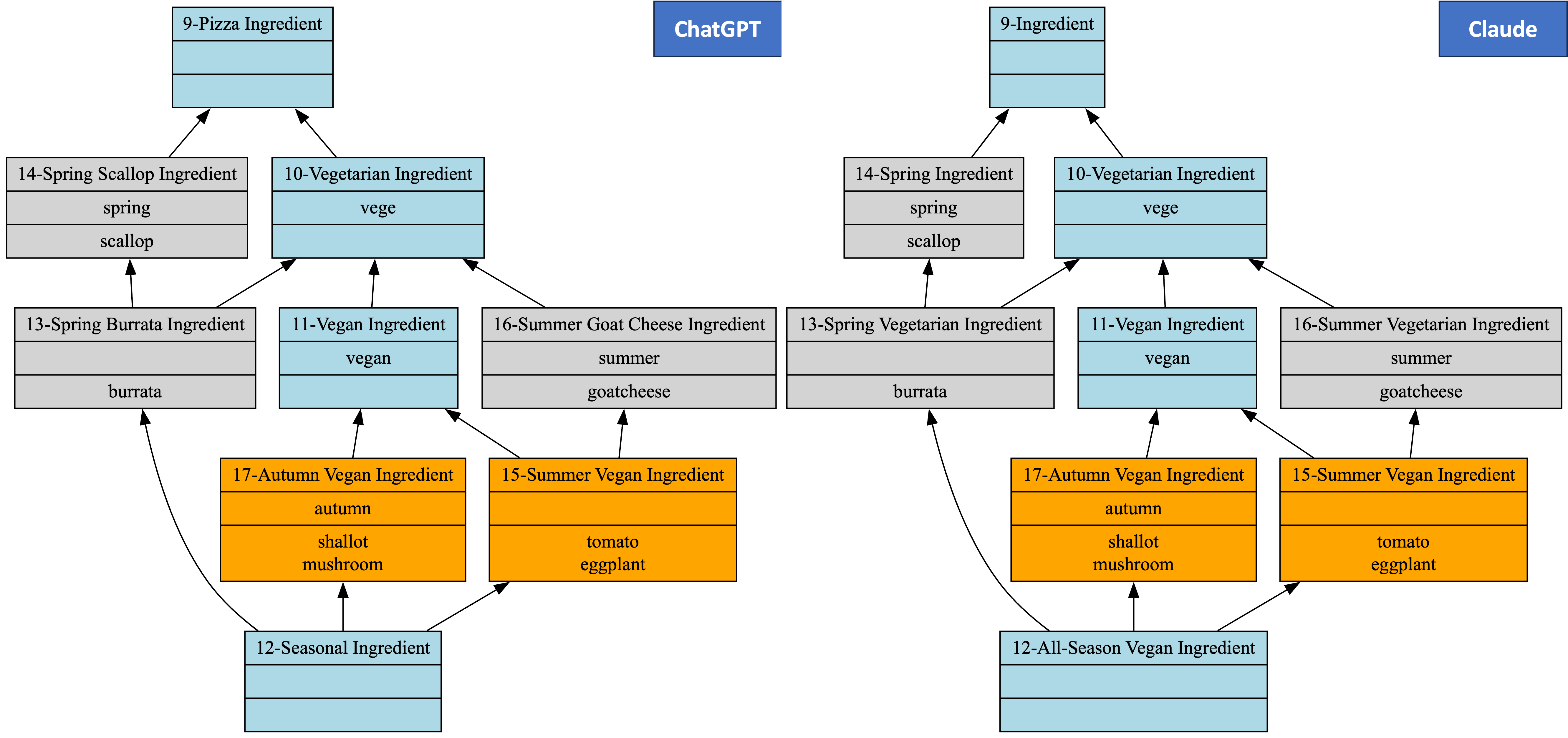}
  \caption{Ingredient lattices from Trials 14 (left-hand side) and 15 (right-hand side). The number preceding the concept name corresponds to the concept ID in Fig \ref{fig_pizzanora}.
  }
  \label{fig_IngredientsChatGPTClaude}
\end{figure}

Table \ref{table_C35} and \ref{table_C36} present the suggested names for these two concepts in all configurations.
From this analysis, two “best” naming proposals for each concept are then selected for the comprehensive lattice naming analysis.  
Figures \ref{fig_PizzeriasChatGPTClaude}, \ref{fig_PizzasChatGPTClaude} and \ref{fig_IngredientsChatGPTClaude} present the naming of the concepts for the three lattices of these two selected  trials.  

\begin{table}[htb]
\renewcommand{\arraystretch}{1.2}
\setlength{\tabcolsep}{0.9pt}
\fontsize{6.5pt}{7pt}\selectfont
\centering
\caption{Variability in names for Concept CPizzer35}
\label{table_C35}
\begin{tabular}{|c|c|c|c|c|l|l|}
\hline
\textbf{Trial} &\textbf{LLM} & \textbf{Concept-} & \textbf{Rel-} & \textbf{\#} & \textbf{Chosen name} & \textbf{Other proposed names} \\ 
\textbf{Id} & & \textbf{Graph} & \textbf{Attrib} & \textbf{Run} &  & \\ \hline
 
1 & ChatGPT & One & Rai & 1 & Vegan Pesto Pizzeria & Summer Vegan Pesto Pizzeria; Veg Summer  \\ 
&  &  &  &  &  &  Pesto Pizzeria; Plant-Based Pesto Pizzeria \\ \hline

2 & ChatGPT & One & Rai & 2 & Vegan Pesto Pizzeria & Summer Vegan Pesto Pizzeria; Vegetal  \\ 
 & &  &  &  &  &  PestoPizzeria; Vegan Summer Pesto  \\  
 & &  &  &  &  &  Pizzeria \\\hline

3 & Claude & One & Rai & 1 & Vegan Pesto Pizzeria & Seasonal Vegan Pesto Pizzeria, Exclusive  \\ 
 & &  &  &  &  &  Vegan Pesto Server, Pure Pesto Vegan  \\ 
 & &  &  &  &  &  Pizzeria \\ \hline

4 & Claude & One & Rai & 2 & Vegan Pesto Pizzeria & Vegan Summer Pesto Pizzeria, Pesto Vegan  \\ 
 & &  &  &  &  &  Pizzeria, Exclusive Pesto Pizzeria \\ \hline

9 & ChatGPT & Several & Rai & 1 & Summer Vegan Pesto & Summer Vegan Pesto Pizzeria; Pesto Pizza \\
 & &  &  & & Pizzeria & Pizzeria; Vegan Pesto Pizzeria \\ \hline

10 & ChatGPT & Several & Rai & 2 & Summer Vegan Pesto  & Summer Pesto Pizzerias; Pesto Pizza \\ 
 & &  &  &  & Pizzerias &  Pizzerias \\ \hline

11 & Claude & Several & Rai & 1 & Pesto Vegan Summer  & Pesto Summer Pizzeria; Vegan Pesto \\ 
 & &  &  &  & Pizzeria & Pizzeria \\ \hline

12 & Claude & Several & Rai & 2 & Pesto Vegan Pizzeria & Vegan Pesto Pizza Pizzeria; Summer Pesto \\ 
 & &  &  &  &  &  Pizzeria \\ \hline

13 & ChatGPT & Several & Raw & 1 & Pesto Summer Veg & Pesto Veg Pizza Place; Summer Pesto Pizza  \\ 
 & &  &  & & Pizzeria &  Shop \\ \hline

14 & ChatGPT & Several & Raw & 2 & Pesto Pizza Pizzeria & Pesto Style Pizzeria; Pesto Pizza Shop \\ \hline

15 & Claude & Several & Raw & 1 & Pesto Pizza Pizzeria & Pesto Vegan Pizzeria; Summer Pesto \\ 
 &  &  &  &  &  &  Pizzeria \\ \hline

16 & Claude & Several & Raw & 2 & Pesto Summer Pizzeria & Summer Pesto Pizzeria; Violet-Style \\ 
 &  &  &  &  &  &  Pizzeria \\ \hline
\end{tabular}
\end{table}

\begin{table}[htb]
\renewcommand{\arraystretch}{1.2}
\setlength{\tabcolsep}{0.9pt}
\fontsize{6.5pt}{7pt}\selectfont
\centering
\caption{Variability in names for Concept CPizzer36}
\label{table_C36}
\begin{tabular}{|c|c|c|c|c|l|l|}
\hline
\textbf{Trial} & \textbf{LLM} & \textbf{Concept-} & \textbf{Rel-} & \textbf{\#} & \textbf{Chosen name} & \textbf{Other proposed names} \\ 
\textbf{Id} & & \textbf{Graph} & \textbf{Attrib} & \textbf{Run} & & \\ \hline
 
5 & ChatGPT & One & Rai & 1 & Vegan Pizza Pizzeria & Vegan-Only Pizza Place;Vegan Ingredient \\
&  &  &  &  & & Pizzeria; Strict Vegan Pizza Pizzeria \\ \hline

6 & ChatGPT & One & Rai & 2 & Vegan Pizza Pizzeria & Vegan-Ingredient Pizza Pizzeria, Strict  \\
 & &  &  &  & &  Vegan Pizza Pizzeria, Only-Vegan Pizza  \\
 & &  &  &  & &  Pizzeria \\ \hline

7 & Claude & One & Rai & 1 & Vegan-Only Pizzeria & Exclusive Vegan Pizza Pizzeria, Vegan \\
 & &  &  & & &  Pizzeria, Purely Vegan Pizza Pizzeria \\ \hline

8 & Claude & One & Rai & 2 & Strict Vegan Pizzeria & Vegan-Only Pizza Pizzeria, Exclusive Vegan \\ 
 & &  &  &  &  & Pizza Pizzeria, All-Vegan Pizza Pizzeria \\ \hline

9 & ChatGPT & Several & Rai & 1 & Vegan Pizza Pizzeria & Vegan Pizza Place; Plant-Based Pizza  \\ 
 &  &  &  &  &  & Pizzeria \\ \hline

10 & ChatGPT & Several & Rai & 2 & Vegan Pizza Pizzerias & Vegan Pizza Places; Pizzerias Serving Vegan  \\ 
 &  &  &  &  &  &  Pizza \\ \hline

11 & Claude & Several & Rai & 1 & Vegan Pizzeria & Plant-Based Pizzeria; Vegan-Friendly \\
 &  &  &  &  &  &  Pizzeria \\ \hline

12 & Claude & Several & Rai & 2 & Vegan Pizzeria & Plant-Based Pizzeria; Strictly Vegan  \\
 &  &  &  &  &  & Pizzeria \\ \hline

13 & ChatGPT & Several & Raw & 1 & Vegan Pizza Pizzeria & Vegan Pizza Restaurant; Plant-Based Pizza \\
 &  &  &  &  &  &  Shop \\ \hline

14 & ChatGPT & Several & Raw & 2 & Vegan Pizza Pizzeria & Vegan Menu Pizzeria; Plant Pizza Pizzeria \\ \hline

15 & Claude & Several & Raw & 1 & Vegan Pizza Pizzeria & Plant-Based Pizzeria; Vegan-Serving  \\
 &  &  &  &  &  &  Pizzeria \\ \hline

16 & Claude & Several & Raw & 2 & Vegan-Serving Pizzeria & Vegan Pizza Pizzeria; Plant-Based Pizzeria \\ \hline

\end{tabular}
\end{table}

\paragraph{Analysis of CPizzer35 and CPizzer36 when concepts are solely named.}

For the four trials naming only concept \texttt{CPizzer35}, the obtained name is “Vegan Pesto Pizzeria" (cf. Table \ref{table_C35}), as this pizzeria serves vegan pizzas and proposes the pesto sauce on their topping. The term “Pesto" is provided by concept \texttt{CPizza22} through the relational attribute, and “Vegan" is inherited from \texttt{CIng11} as \texttt{CIng15} does not introduce any attribute. Unfortunately, this name presents many flaws. “Vegan" is a quality of pizza topping ingredients and “Pesto" is a sauce or an aromat (not present in the lattice dedicated to ingredients and thus ambiguous) and they quite differ in nature. Here, these two adjectives describe the pizzeria, i.e, the restaurant where pizzas are served, in a metonymic relation that introduces a bias. Moreover, the order is misleading. Given as such, one could think that the pizzeria is specialized in pesto (soup? sauce? plant? as the word is polysemous) and the pizzeria is also vegan (a  metonymic shortcut to  “mostly serving vegan dishes"). One could reduce this double metonymy and the ambiguity by introducing the focus for which the adjectives are relevant, here the pizza. When adding “pizza" to the name, a noun which is correctly qualified by the nature of its topping (e.g. “vegan")  and the specificity of its flavor (e.g “pesto"), then ambiguity is quite lessened. 

For concept \texttt{CPizzer36}, ChatGPT suggests “Vegan Pizza Pizzeria" in both runs, and Claude Sonnet “Vegan only Pizzeria" and “Strict Vegan Pizzeria" (cf. Table \ref{table_C36}). Comparing these suggestions, ChatGPT is  better than Claude. ChatGPT introduces the missing object of the metonymic relation, i.e, the pizza, and correctly attributes “Vegan" to the pizza. Thus the name “Vegan Pizza Pizzeria" is quite valid. 
Using “Vegan only" and “Strict vegan", Claude is too restrictive (probably a confusion between $\exists$ and $\forall$ quantifiers in the relation description). The metonymy object is also absent.

\paragraph{Analysis of CPizzer35 and CPizzer36 when all concepts are named} 

The \texttt{Raw} (relational attributes not rewritten) trials behave better than \texttt{Rai} (relational attributes are rewritten) in that they provide the least metonymic name. All names excluding the word “pizza" are metonymic and thus more ambiguous since they could hint at a completely different notion.
\\For concept \texttt{CPizzer35}, in some trials, the suggested concept name combines the season and the vegetables belonging to the topping ingredients in a totally misleading order, such as “Pesto Vegan Summer Pizzeria" (Rai Claude Sonnet, Trial 11 in Table \ref{table_C35}).

The least questionable name is given by Trials 14 and 15, 
 “Pesto Pizza Pizzeria", where the word pizza is present. However, the influence of primary vs secondary attributes is not reflected here. Pesto as a sauce or a flavor is less important than the ingredient nature (Vegan). 
\\According to the results tables, it seems that the number of runs has an effect, since Trial 16 (vs Trial 15) 
suggests the name “Pesto Summer Pizzeria", dropping thus the true object to which “Pesto" and “Summer" could be better attached, and introducing a completely different understanding of the concept (the pizzeria could be seen as a summer restaurant, probably not open in other times).  A possible name could be “Pesto Summer Veg Pizzeria". In this name “Pizza" is lacking, but “Summer" appropriately applies to vegetables. However, we also lack the fact that pizzas could be vegan friendly, since toppings with summer vegetables could also include non vegan ingredients. 
Thus, in general, for concept \texttt{CPizzer35} the names are not quite satisfactory. 
A consistent name may be
“Vegan Pesto Pizza Pizzeria" (adding “Pizza" to the outcome of trials 1 to 4). The seasonal aspect is not considered as mandatory in the title, and the “Pesto" sauce (or flavor) is the differentiating aspect between \texttt{CPizzer35} and its parent \texttt{CPizzer36}. 
\\ For concept \texttt{CPizzer36}, several trials provide an acceptable name, i.e, the ones including the word “Pizza". In fact, this concept describes a pizzeria that serves vegan pizzas. Trials 14 and 15  
both give “Vegan Pizza Pizzeria" agreeing thus with the isolated concept naming. An interesting result is given by Trial 15  
with “Vegan serving Pizzeria" in its proposal list. The object is missing but the relation is present, thus lessening the effect of metonymy. It seems that the afterthoughts of Claude in the second run tend to introduce information that is sometimes less relevant and sometimes broader and possibly more flexible than the most local name. Because they provide informative naming proposals, these results make Trial 14 and Trial 15 the selected results.

\begin{table}[htb]
\renewcommand{\arraystretch}{1.2}
\setlength{\tabcolsep}{0.9pt}
\fontsize{6.5pt}{7pt}\selectfont
\centering
\caption{Variability in the names of concepts chosen by the LLMs for the two selected trials}
\label{table_BestTwo}
\begin{tabular}{|l|c|l|l|}
\hline
\textbf{Lattice} & \textbf{Concept} & \textbf{Chosen name for trial 14} & \textbf{Chosen name for trial 15} \\ \textbf{} & \textbf{Id} & (ChatGPT) & (Claude) \\\hline
Ingredient & 9 & Pizza Ingredient & Ingredient \\ \hline
Ingredient & 10 & Vegetarian Ingredient & Vegetarian Ingredient \\ \hline
Ingredient & 11 & Vegan Ingredient & Vegan Ingredient \\ \hline
Ingredient & 12 & Seasonal Ingredient & All-Season Vegan Ingredient \\ \hline
Ingredient & 13 & Spring Burrata Ingredient & Spring Vegetarian Ingredient \\ \hline
Ingredient & 14 & Spring Scallop Ingredient & Spring Ingredient \\ \hline
Ingredient & 15 & Summer Vegan Ingredient & Summer Vegan Ingredient \\ \hline
Ingredient & 16 & Summer Goat Cheese Ingredient & Summer Vegetarian Ingredient \\ \hline
Ingredient & 17 & Autumn Vegan Ingredient & Autumn Vegan Ingredient \\ \hline
Pizza & 18 & Ingredient Pizza & Pizza \\ \hline
Pizza & 19 & Red Goat Cheese Pizza & Red Summer Vegetarian Pizza \\ \hline
Pizza & 20 & Specialty Sauce Pizza & All-Seasonal Vegan Pizza \\ \hline
Pizza & 21 & BBQ Autumn Pizza & BBQ Vegan Pizza \\ \hline
Pizza & 22 & Pesto Summer Pizza & Pesto Vegan Pizza \\ \hline
Pizza & 23 & White Scallop Pizza & White Spring Pizza \\ \hline
Pizza & 30 & Vegetarian Pizza & Vegetarian Pizza \\ \hline
Pizza & 31 & Vegan Pizza & Vegan Pizza \\ \hline
Pizza & 32 & Summer Vegetarian Pizza & Summer Vegetarian Pizza \\ \hline
Pizzeria & 24 & Multi Pizza Pizzeria & Pizzeria \\ \hline
Pizzeria & 25 & Food Truck Pizzeria & Food-Truck Pizzeria \\ \hline
Pizzeria & 26 & Full Service Pizzeria & Full-Service Pizzeria \\ \hline
Pizzeria & 27 & Takeaway Pizzeria & Takeaway Pizzeria \\ \hline
Pizzeria & 28 & Delivery Pizzeria & Delivery Pizzeria \\ \hline
Pizzeria & 29 & Dine In Pizzeria & Dine-In Pizzeria \\ \hline
Pizzeria & 33 & Red Pizza Pizzeria & Red Pizza Pizzeria \\ \hline
Pizzeria & 34 & BBQ Pizza Pizzeria & BBQ Pizza Pizzeria \\ \hline
Pizzeria & 35 & Pesto Pizza Pizzeria & Pesto Pizza Pizzeria \\ \hline
Pizzeria & 36 & Vegan Pizza Pizzeria & Vegan Pizza Pizzeria \\ \hline
\end{tabular}
\end{table}

\paragraph{Analysis of the name of the  Pizzeria lattice concepts from the two selected trials}

Table  \ref{table_BestTwo} summarizes the chosen name of concepts for all the lattices from Trials 14 and 15.
The obtained results show some interesting tracks, that are here analyzed according to the concept generalization level.
\\When dealing with pizzeria lattices (cf. Fig. \ref{fig_PizzeriasChatGPTClaude}), i.e., concepts \texttt{CPizzer24} to \texttt{CPizzer36}, both LLMs agree when an introduced attribute is in the concept (e.g, “dine-in", “food-truck", etc.). For the top concept \texttt{CPizzer24}, ChatGPT is imaginative with a “Multi Pizza Pizzeria" name, whereas Claude restricts to the essence of the concept itself, naming it “Pizzeria" solely. In fact, in such situation, Claude strategy is the best, because the higher the concept is, the more general name the better. “Multi-pizza" looks like an obvious statement for a pizzeria. 
\\For concepts containing 
relational attributes only, both trials agree on the name pattern, that is, the qualified relation argument name followed by “Pizzeria". So Claude and ChatGPT (but for one) provide the best possible names in the local context. Here “Pesto Pizza Pizzeria", the name retained for \texttt{CPizzer35}, appears as a child of “Vegan Pizza Pizzeria". This name does not include the parent features since the naming considers inheritance as an implicit factor and avoids redundancy. This points at the importance of “relativity" in names. When concepts are isolated, one tends to prefer the most explicit names while avoiding redundancy. When concepts are looked at in the model context, the name is seen in the context of its parent concept and thus  the elision phenomenon is at work. However, it is better to impose the non-redundant non-ambiguous name of “Pesto Vegan Pizza Pizzeria". 
For \texttt{CPizzer33} and \texttt{CPizzer34}, which also have no proper attributes but a relational one, it is interesting to see how the name was chosen based on this attribute. “Red Pizza" qualifies the pizza \texttt{CPizza19} which in fact is named “Red Summer Vegetarian Pizza" by Claude and “Red Goat Cheese Pizza" by ChatGPT.
The transmitted name is reduced to the proper attribute of the relation argument (here “Red") as well as its basic category “Pizza", omitting the distinctiveness. The strategy obeys to the limit of four units, but also, the picking is quite sensible. A particular attention has to be given to the pizzeria lattice bottom concept, \texttt{CPizzer26}. Both LLMs provide the same name. But in the list of proposals “All mode Pizzeria" (Claude) or “Multi-service Pizzeria" (ChatGPT) are also acceptable names. 

\paragraph{Analysis of the Pizza lattice concept names from the two selected trials} When dealing with pizza lattices (cf. Fig. \ref{fig_PizzasChatGPTClaude}), i.e, concepts \texttt{CPizza18}  to \texttt{CPizza23} and \texttt{CPizza30} to \texttt{CPizza32} in Table  \ref{table_BestTwo}, differences in names begin to appear.
For the three last ones, \texttt{CPizza30} to \texttt{CPizza32}, the two LLMs agree. Given the constraints in the model, names are considered valid, although some questions are raised about the intrinsic consistency of the model.  
\\For concepts \texttt{CPizza18} to \texttt{CPizza30}, differences are interesting and denote local trends in the LLMs. ChatGPT tends to state the obvious and sometimes provides an absurd denomination (“Ingredient Pizza" for \texttt{CPizza18}) whereas Claude abides by a consistent strategy in maintaining generality for uppermost concepts (“Pizza" is just enough for \texttt{CPizza18}). Same behavior is observed for \texttt{CPizzeria24} (the top concept of Pizzeria lattices).
\\For concepts \texttt{CPizza21} to \texttt{CPizza23}, Claude strategy is quite clear in its preference. It  first uses the introduced attribute, then the relational attribute. The first adjective is given by the local attribute, e.g, “BBQ", “Pesto", “White". Then  it moves to the introduced attribute of the relation argument parent if it exists e.g, “Vegan" (\texttt{CPizza21}, \texttt{CPizza22}), or the relation argument introduced attribute if it exists and the parent is empty (case of “Spring" in \texttt{CPizza23}). 
Whereas ChatGPT, beginning with the same strategy, prefers the introduced attribute of the relation argument if it exists (e.g “Autumn" in \texttt{CPizza21}). If not, it chooses among parents the one that is not related to a parent of the concept: here, it prefers “Summer" to “Vegan" in \texttt{CPizza22} since “Vegan" is related to the parent \texttt{CPizza31} of \texttt{CPizza21}. 
The question is thus why it does not adopt the strategy used for \texttt{CPizza23}. 
It begins with “White", then prefers the object “Scallop", that belongs to “Spring" ingredients, to the relation argument introduced attribute, “Spring". 
\\For concept \texttt{CPizza19}, named “Red Goat Cheese Pizza" by ChatGPT and “Red Summer Vegetarian Pizza" by Claude, the same strategy as in \texttt{CPizza23} applies. An explanation could be the following: when an object is present, it is more specific than the concept introduced attribute shared by all objects. The rationale behind naming is clear in that sense. 
\\The difference in the name of the bottom concept \texttt{CPizza20} perfectly illustrates this divergence in strategy. 
The name chosen by ChatGPT and Claude is respectively “Specialty Sauce Pizza" and “All-seasonal Vegan Pizza". The rationale is very interesting. Claude privileges the relational argument names and takes the most general among them. Conversely, ChatGPT looks first for objects and then for attributes located in parent concepts. 
A consistent behavior then emerges from both name proposals: ChatGPT relies on intent while Claude favors relations arguments and constraints. 
\\The analysis of the concepts located within the lattice exhibits two main results. (i) There is a divergence in the strategy to choose a word for the name units  that are in the middle. Unit one is generally the same and consists in the introduced attribute of the concept and the last unit corresponds to the generic parent name. For the units in the middle, the choice depends on the importance of the attributes and objects in the relation argument. ChatGPT privileges specificity and objects (intent) whereas Claude stands by more generality, relations constraints, and attributes.
(ii) The differences in naming point at some questionable aspects in the modeling of the dataset, i.e., the relational context family as presented in Table \ref{table_pizzas}. The name of parent concepts, such as “Toppings" and “Sauce", which could generalize most of the attributes and objects, are omitted. Their use would have introduced a more “relation oriented" data model; naming strategies would therefore be better adapted. Some hints about such data modeling and strategies are suggested in the list of proposed names given by the LLMs, e.g., “Specialty Sauce" in the bottom concept \texttt{CPizza20} and “All-season" in the ingredient bottom concept \texttt{CIng12}. 

\paragraph{Analysis of the Ingredient lattice concepts names from the two selected trials}
The ingredient lattices from the two selected trials are presented in Fig. 
\ref{fig_IngredientsChatGPTClaude}. The ingredient part of the data model looks like the most sensitive to modeling shortcomings. 
\\ \texttt{CIng9}, the top lattice concept, is named “Ingredient" by Claude and “Pizza Ingredient" by ChatGPT. Here, the term “Ingredient" is too vague, too polysemous. When asked, ChatGPT provides a list of 20 different usages of the word “ingredient" in several contexts, even within a food context. In the data model context, “Ingredient" would be enough. However, when one looks at it, it shows that most addressed ingredients are those of the topping. It is in one of the ChatGPT proposed names “Pizza topping ingredient". Here the specificity looked for by ChatGPT is quite fruitful, unlike other discussed cases. Applying this name to \texttt{CIng9} will help adding other ingredients that might belong to the crust or to the sauce. If “Pizza topping", “Pizza crust", and “Pizza sauce" are present in the data model, then “topping ingredient" would be enough for \texttt{CIng9}.  
\\ \texttt{CIng10} and \texttt{CIng11} are alike and correspond to the best choices in the context. \texttt{CIng12} is a bottom concept as pointed at in the previous paragraph. Here, it corresponds to the concept that relates ingredients to seasons, those specific and those not specific. Claude is more associative, when proposing “all seasons", whereas ChatGPT insists on the seasonal aspect. Claude adds “vegan", which is a non-relevant adjective. The rationale is the following: “Inherits vegan + all three season attributes (spring, summer, autumn) with an empty extent, a bottom concept representing an impossible combination. “All-Season" signals the exhaustive seasonal constraint causing the empty extent. It should not inherit ``vegan'' since it also inherits “vegetarian" from \texttt{CIng13}, which object is “burrata" (cheese), a non vegan ingredient. In that sense, the empty extent is also constrained by the vegan and non vegan combination. 
\\ \texttt{CIng13} to \texttt{CIng17} privilege the specific season in which the ingredients are found, with a total agreement between both LLMs in \texttt{CIng11}, \texttt{CIng15}, and \texttt{CIng17}. The differences in the other concepts names originated from the LLMs favored strategies: Claude preferring parents proper attributes, and Chatgpt relying on local objects distinguished in the concept, as a second thought item for choosing names. 

\paragraph{Summarizing the qualitative analysis}
In conclusion, agreement between both LLMs is mostly apparent when concepts present proper attributes and relations. The produced names are  human-acceptable. Differences may show up when attributes are not present and/or  when objects are listed. 
In top concepts, Claude is more general in a mainstream acceptability, and ChatGPT is sometimes absurd, obvious and sometimes quite striking. In bottom concepts, variability is high and sensitive to the LLM strategy.  In middle level concepts, when relational attributes are the highlighted elements, both LLMs seem to agree, and the provided names are human-acceptable. When attributes and objects are the only differentiating items, variability in naming depends on the LLM underlying preferences. Claude confirms its generality preference and indicates it in the rationale sentences provided in the table, while ChatGPT chooses specificity, sometimes relevantly and sometimes not, focusing on the intent highlight detected in the concept. Objects are thus present in the name, especially if they are unique in the concept. 

\subsection{Quantitative analysis}
\label{Quantitative_analysis}

\begin{table*}[htb]
\renewcommand{\arraystretch}{1.2}
\setlength{\tabcolsep}{0.9pt}
\centering
\caption{Assessment of the LLMs common names for concepts from trials 14 and 15 by Human judges. 
Score meaning: 1, 0.5, 0, and -0.5 indicate that the chosen name is accepted, a name other than the one chosen but present in the list suggested by the LLM is accepted, the name is not accepted, and an alternative name is suggested, respectively.}
\label{llms_Agree}
\fontsize{6.5pt}{7pt}\selectfont
\begin{tabular}{|c|c|l|cccccc|l|}
\hline
\textbf{Lattice} & \textbf{Concept} & \multicolumn{1}{c|}{\textbf{Concept name}} & \multicolumn{6}{c|}{\textbf{Scores}} & \textbf{Suggested names} \\ 
 &  \textbf{ID} & \multicolumn{1}{c|}{} & \multicolumn{1}{c|}{\textbf{J1}} & \multicolumn{1}{c|}{\textbf{J2}} & \multicolumn{1}{c|}{\textbf{J3}} & \multicolumn{1}{c|}{\textbf{J4}} & \multicolumn{1}{c|}{\textbf{J5}} & \textbf{Total} &  \\ \hline
Ingredient & 10 & Vegetarian Ingredient & \multicolumn{1}{c|}{1} & \multicolumn{1}{c|}{1} & \multicolumn{1}{c|}{1} & \multicolumn{1}{c|}{1} & \multicolumn{1}{c|}{1} & 5 & \\ \hline
Ingredient & 11 & Vegan Ingredient & \multicolumn{1}{c|}{1} & \multicolumn{1}{c|}{1} & \multicolumn{1}{c|}{1} & \multicolumn{1}{c|}{1} & \multicolumn{1}{c|}{1} & 5 & \\ \hline
Ingredient & 15 & Summer Vegan Ingredient & \multicolumn{1}{c|}{1} & \multicolumn{1}{c|}{1} & \multicolumn{1}{c|}{1} & \multicolumn{1}{c|}{1} & \multicolumn{1}{c|}{1} & 5 & \\ \hline
Ingredient & 17 & Autumn Vegan Ingredient & \multicolumn{1}{c|}{1} & \multicolumn{1}{c|}{1} & \multicolumn{1}{c|}{1} & \multicolumn{1}{c|}{1} & \multicolumn{1}{c|}{1} & 5 & \\ \hline
Pizza & 30 & Vegetarian Pizza & \multicolumn{1}{c|}{1} & \multicolumn{1}{c|}{1} & \multicolumn{1}{c|}{1} & \multicolumn{1}{c|}{1} & \multicolumn{1}{c|}{1} & 5 & \\ \hline
Pizza & 31 & Vegan Pizza & \multicolumn{1}{c|}{1} & \multicolumn{1}{c|}{1} & \multicolumn{1}{c|}{1} & \multicolumn{1}{c|}{1} & \multicolumn{1}{c|}{1} & 5 & \\ \hline
Pizza & 32 & Summer Vegetarian Pizza & \multicolumn{1}{c|}{1} & \multicolumn{1}{c|}{1} & \multicolumn{1}{c|}{1} & \multicolumn{1}{c|}{1} & \multicolumn{1}{c|}{1} & 5 & \\ \hline
Pizzeria & 25 & Food Truck Pizzeria & \multicolumn{1}{c|}{1} & \multicolumn{1}{c|}{1} & \multicolumn{1}{c|}{1} & \multicolumn{1}{c|}{1} & \multicolumn{1}{c|}{1} & 5 & \\ \hline

Pizzeria & 26 & Full Service Pizzeria & \multicolumn{1}{c|}{0} & \multicolumn{1}{c|}{0} & \multicolumn{1}{c|}{-0.5} & \multicolumn{1}{c|}{0} & \multicolumn{1}{c|}{-0.5} & -1 & All-mode pizzeria; All Pizzeria; \\
 &  &  & \multicolumn{1}{c|}{} & \multicolumn{1}{c|}{} & \multicolumn{1}{c|}{} & \multicolumn{1}{c|}{} & \multicolumn{1}{c|}{} &  & All service Pizzeria; Pizzeria \\ 
 &  &  & \multicolumn{1}{c|}{} & \multicolumn{1}{c|}{} & \multicolumn{1}{c|}{} & \multicolumn{1}{c|}{} & \multicolumn{1}{c|}{} &  &  Service Type \\ \hline

Pizzeria & 27 & Takeaway Pizzeria & \multicolumn{1}{c|}{1} & \multicolumn{1}{c|}{1} & \multicolumn{1}{c|}{1} & \multicolumn{1}{c|}{1} & \multicolumn{1}{c|}{1} & 5 & \\ \hline
Pizzeria & 28 & Delivery Pizzeria & \multicolumn{1}{c|}{1} & \multicolumn{1}{c|}{1} & \multicolumn{1}{c|}{1} & \multicolumn{1}{c|}{1} & \multicolumn{1}{c|}{1} & 5 & \\ \hline
Pizzeria & 29 & Dine In Pizzeria & \multicolumn{1}{c|}{1} & \multicolumn{1}{c|}{1} & \multicolumn{1}{c|}{1} & \multicolumn{1}{c|}{1} & \multicolumn{1}{c|}{1} & 5 & \\ \hline

Pizzeria & 33 & Red Pizza Pizzeria & \multicolumn{1}{c|}{-0.5} & \multicolumn{1}{c|}{-0.5} & \multicolumn{1}{c|}{-0.5} & \multicolumn{1}{c|}{-0.5} & \multicolumn{1}{c|}{1} & -1 & Red-pizza serving pizzeria; \\  & & & \multicolumn{1}{c|}{} & \multicolumn{1}{c|}{} & \multicolumn{1}{c|}{} & \multicolumn{1}{c|}{} & \multicolumn{1}{c|}{} &  & Pizzeria serving Red Pizza \\ \hline

Pizzeria & 34 & BBQ Pizza Pizzeria & \multicolumn{1}{c|}{-0.5} & \multicolumn{1}{c|}{-0.5} & \multicolumn{1}{c|}{-0.5} & \multicolumn{1}{c|}{-0.5} & \multicolumn{1}{c|}{1} & -1 & BBQ-pizza serving pizzeria; \\
 & & & \multicolumn{1}{c|}{} & \multicolumn{1}{c|}{} & \multicolumn{1}{c|}{} & \multicolumn{1}{c|}{} & \multicolumn{1}{c|}{} &  & Pizzeria serving BBQ Pizza \\ \hline

Pizzeria & 35 & Pesto Pizza Pizzeria & \multicolumn{1}{c|}{-0.5} & \multicolumn{1}{c|}{0.5} & \multicolumn{1}{c|}{-0.5} & \multicolumn{1}{c|}{-0.5} & \multicolumn{1}{c|}{-0.5} & -1.5 & Pesto-serving pizza pizzeria;  \\
 & & & \multicolumn{1}{c|}{} & \multicolumn{1}{c|}{} & \multicolumn{1}{c|}{} & \multicolumn{1}{c|}{} & \multicolumn{1}{c|}{} &  & Pizzeria serving Pesto Pizza;  \\ 
 & & & \multicolumn{1}{c|}{} & \multicolumn{1}{c|}{} & \multicolumn{1}{c|}{} & \multicolumn{1}{c|}{} & \multicolumn{1}{c|}{} &  & Vegan pesto pizza pizzeria \\ \hline

Pizzeria & 36 & Vegan Pizza Pizzeria & \multicolumn{1}{c|}{0.5} & \multicolumn{1}{c|}{-0.5} & \multicolumn{1}{c|}{-0.5} & \multicolumn{1}{c|}{-0.5} & \multicolumn{1}{c|}{1} & 0 & Vegan-serving pizza pizzeria; \\
 &  &  & \multicolumn{1}{c|}{} & \multicolumn{1}{c|}{} & \multicolumn{1}{c|}{} & \multicolumn{1}{c|}{} & \multicolumn{1}{c|}{} & &  Pizzeria serving Vegan Pizza \\\hline
\end{tabular}
\end{table*}

\begin{table}[htb]
\renewcommand{\arraystretch}{1.2}
\setlength{\tabcolsep}{0.9pt}
\fontsize{6.5pt}{7pt}\selectfont
\centering
\caption{Evaluation of the alternatives names proposed by the LLMs for the concepts from trials 14 and 15 by Human Judges. Score meaning: 1 is accepted (0 to the other), 0.5 if a name of the proposal list is  accepted (either 0.5 or 0 to the other), 0 to both  if neither is accepted, -0.5 (either -0.5 or 0 to the other) if an  alternative name outside the list is suggested (0 to the closest)}
\label{llms_Disagree}
\begin{tabular}{|l|c|c|cccccc|c|cccccc|}
\hline
\textbf{Lattice} & \textbf{Cpt} & \textbf{Concept name} & \multicolumn{6}{c|}{\textbf{Scores}} & \textbf{Concept name} & \multicolumn{6}{c|}{\textbf{Scores}} \\ 
 & \textbf{ID} & \textbf{Trial 15} & \multicolumn{1}{c|}{\textbf{J1}} & \multicolumn{1}{c|}{\textbf{J2}} & \multicolumn{1}{c|}{\textbf{J3}} & \multicolumn{1}{c|}{\textbf{J4}} & \multicolumn{1}{c|}{\textbf{J5}} & \textbf{Sum} & \textbf{Trial 14} & \multicolumn{1}{c|}{\textbf{J1}} & \multicolumn{1}{c|}{\textbf{J2}} & \multicolumn{1}{c|}{\textbf{J3}} & \multicolumn{1}{c|}{\textbf{J4}} & \multicolumn{1}{c|}{\textbf{J5}} & \textbf{Sum} \\ \hline
Ingredient & 9 & Ingredient & \multicolumn{1}{c|}{0} & \multicolumn{1}{c|}{0} & \multicolumn{1}{c|}{-0.5} & \multicolumn{1}{c|}{0} & \multicolumn{1}{c|}{0} & -0.5 & pizza ingredient & \multicolumn{1}{c|}{0.5} & \multicolumn{1}{c|}{0.5} & \multicolumn{1}{c|}{-0.5} & \multicolumn{1}{c|}{1} & \multicolumn{1}{c|}{0.5} & 2 \\ \hline

Ingredient & 12 & All-Season Vegan & \multicolumn{1}{c|}{1} & \multicolumn{1}{c|}{1} & \multicolumn{1}{c|}{0} & \multicolumn{1}{c|}{0.5} & \multicolumn{1}{c|}{0.5} & 3 & seasonal ingredient & \multicolumn{1}{c|}{0} & \multicolumn{1}{c|}{0} & \multicolumn{1}{c|}{1} & \multicolumn{1}{c|}{0.5} & \multicolumn{1}{c|}{0.5} & 2 \\ 
 &  & Ingredient & \multicolumn{1}{c|}{} & \multicolumn{1}{c|}{} & \multicolumn{1}{c|}{} & \multicolumn{1}{c|}{} & \multicolumn{1}{c|}{} & &  & \multicolumn{1}{c|}{} & \multicolumn{1}{c|}{} & \multicolumn{1}{c|}{} & \multicolumn{1}{c|}{} & \multicolumn{1}{c|}{} &  \\ \hline

Ingredient & 13 & Spring Vegetarian  & \multicolumn{1}{c|}{1} & \multicolumn{1}{c|}{1} & \multicolumn{1}{c|}{1} & \multicolumn{1}{c|}{1} & \multicolumn{1}{c|}{1} & 5 & spring burrata & \multicolumn{1}{c|}{0} & \multicolumn{1}{c|}{0} & \multicolumn{1}{c|}{0} & \multicolumn{1}{c|}{0} & \multicolumn{1}{c|}{0} & 0 \\ 
 &  & Ingredient & \multicolumn{1}{c|}{} & \multicolumn{1}{c|}{} & \multicolumn{1}{c|}{} & \multicolumn{1}{c|}{} & \multicolumn{1}{c|}{} &  & ingredient & \multicolumn{1}{c|}{} & \multicolumn{1}{c|}{} & \multicolumn{1}{c|}{} & \multicolumn{1}{c|}{} & \multicolumn{1}{c|}{} &  \\ \hline

Ingredient & 14 & Spring Ingredient & \multicolumn{1}{c|}{1} & \multicolumn{1}{c|}{1} & \multicolumn{1}{c|}{1} & \multicolumn{1}{c|}{1} & \multicolumn{1}{c|}{1} & 5 & spring scallop  & \multicolumn{1}{c|}{0} & \multicolumn{1}{c|}{0} & \multicolumn{1}{c|}{0} & \multicolumn{1}{c|}{0} & \multicolumn{1}{c|}{0} & 0 \\
 &  &  & \multicolumn{1}{c|}{} & \multicolumn{1}{c|}{} & \multicolumn{1}{c|}{} & \multicolumn{1}{c|}{} & \multicolumn{1}{c|}{} & 5 & ingredient & \multicolumn{1}{c|}{} & \multicolumn{1}{c|}{} & \multicolumn{1}{c|}{} & \multicolumn{1}{c|}{} & \multicolumn{1}{c|}{} &  \\ \hline

Ingredient & 16 & Summer Vegetarian  & \multicolumn{1}{c|}{1} & \multicolumn{1}{c|}{1} & \multicolumn{1}{c|}{1} & \multicolumn{1}{c|}{1} & \multicolumn{1}{c|}{1} & 5 & Summer Goat & \multicolumn{1}{c|}{0} & \multicolumn{1}{c|}{0} & \multicolumn{1}{c|}{0} & \multicolumn{1}{c|}{0} & \multicolumn{1}{c|}{0} & 0 \\ 
 &  & Ingredient & \multicolumn{1}{c|}{} & \multicolumn{1}{c|}{} & \multicolumn{1}{c|}{} & \multicolumn{1}{c|}{} & \multicolumn{1}{c|}{} &  & Cheese Ingredient & \multicolumn{1}{c|}{} & \multicolumn{1}{c|}{} & \multicolumn{1}{c|}{} & \multicolumn{1}{c|}{} & \multicolumn{1}{c|}{} &  \\ \hline

Pizza & 18 & Pizza & \multicolumn{1}{c|}{1} & \multicolumn{1}{c|}{1} & \multicolumn{1}{c|}{-0.5} & \multicolumn{1}{c|}{1} & \multicolumn{1}{c|}{1} & 3.5 & Ingredient Pizza & \multicolumn{1}{c|}{0} & \multicolumn{1}{c|}{0} & \multicolumn{1}{c|}{-0.5} & \multicolumn{1}{c|}{0} & \multicolumn{1}{c|}{0} & -0.5 \\ \hline

Pizza & 19 & Red Summer & \multicolumn{1}{c|}{1} & \multicolumn{1}{c|}{1} & \multicolumn{1}{c|}{-0.5} & \multicolumn{1}{c|}{0.5} & \multicolumn{1}{c|}{1} & 3 & Red Goat Cheese  & \multicolumn{1}{c|}{0} & \multicolumn{1}{c|}{0} & \multicolumn{1}{c|}{-0.5} & \multicolumn{1}{c|}{0.5} & \multicolumn{1}{c|}{0} & 0 \\
 &  & Vegetarian Pizza & \multicolumn{1}{c|}{} & \multicolumn{1}{c|}{} & \multicolumn{1}{c|}{} & \multicolumn{1}{c|}{} & \multicolumn{1}{c|}{} &  & Pizza & \multicolumn{1}{c|}{} & \multicolumn{1}{c|}{} & \multicolumn{1}{c|}{} & \multicolumn{1}{c|}{} & \multicolumn{1}{c|}{} &  \\ \hline

Pizza & 20 & All-Seasonal & \multicolumn{1}{c|}{0} & \multicolumn{1}{c|}{1} & \multicolumn{1}{c|}{1} & \multicolumn{1}{c|}{1} & \multicolumn{1}{c|}{1} & 4 & Specialty Sauce & \multicolumn{1}{c|}{0} & \multicolumn{1}{c|}{0} & \multicolumn{1}{c|}{0} & \multicolumn{1}{c|}{0} & \multicolumn{1}{c|}{0} & 0 \\
 &  & Vegan Pizza & \multicolumn{1}{c|}{} & \multicolumn{1}{c|}{} & \multicolumn{1}{c|}{} & \multicolumn{1}{c|}{} & \multicolumn{1}{c|}{} &  & Pizza & \multicolumn{1}{c|}{} & \multicolumn{1}{c|}{} & \multicolumn{1}{c|}{} & \multicolumn{1}{c|}{} & \multicolumn{1}{c|}{} &  \\ \hline

Pizza & 21 & BBQ Vegan & \multicolumn{1}{c|}{0.5} & \multicolumn{1}{c|}{0.5} & \multicolumn{1}{c|}{1} & \multicolumn{1}{c|}{1} & \multicolumn{1}{c|}{1} & 4 & BBQ Autumn & \multicolumn{1}{c|}{0} & \multicolumn{1}{c|}{0} & \multicolumn{1}{c|}{0} & \multicolumn{1}{c|}{-0.5} & \multicolumn{1}{c|}{0} & -0.5 \\ 
 &  & Pizza & \multicolumn{1}{c|}{} & \multicolumn{1}{c|}{} & \multicolumn{1}{c|}{} & \multicolumn{1}{c|}{} & \multicolumn{1}{c|}{} &  & Pizza & \multicolumn{1}{c|}{} & \multicolumn{1}{c|}{} & \multicolumn{1}{c|}{} & \multicolumn{1}{c|}{} & \multicolumn{1}{c|}{} &  \\ \hline

Pizza & 22 & Pesto Vegan & \multicolumn{1}{c|}{0.5} & \multicolumn{1}{c|}{0.5} & \multicolumn{1}{c|}{1} & \multicolumn{1}{c|}{0} & \multicolumn{1}{c|}{0.5} & 2.5 & Pesto Summer  & \multicolumn{1}{c|}{0} & \multicolumn{1}{c|}{0} & \multicolumn{1}{c|}{0} & \multicolumn{1}{c|}{0.5} & \multicolumn{1}{c|}{0.5} & 1 \\
 &  & Pizza & \multicolumn{1}{c|}{} & \multicolumn{1}{c|}{} & \multicolumn{1}{c|}{} & \multicolumn{1}{c|}{} & \multicolumn{1}{c|}{} &  & Pizza & \multicolumn{1}{c|}{} & \multicolumn{1}{c|}{} & \multicolumn{1}{c|}{} & \multicolumn{1}{c|}{} & \multicolumn{1}{c|}{} &  \\ \hline

Pizza & 23 & White Spring & \multicolumn{1}{c|}{1} & \multicolumn{1}{c|}{0} & \multicolumn{1}{c|}{0} & \multicolumn{1}{c|}{-0.5} & \multicolumn{1}{c|}{-0.5} & 0 & White Scallop & \multicolumn{1}{c|}{0} & \multicolumn{1}{c|}{1} & \multicolumn{1}{c|}{1} & \multicolumn{1}{c|}{0} & \multicolumn{1}{c|}{-0.5} & 1.5 \\ 
 &  & Pizza & \multicolumn{1}{c|}{} & \multicolumn{1}{c|}{} & \multicolumn{1}{c|}{} & \multicolumn{1}{c|}{} & \multicolumn{1}{c|}{} &  & Pizza & \multicolumn{1}{c|}{} & \multicolumn{1}{c|}{} & \multicolumn{1}{c|}{} & \multicolumn{1}{c|}{} & \multicolumn{1}{c|}{} &  \\ \hline

Pizzeria & 24 & Pizzeria & \multicolumn{1}{c|}{1} & \multicolumn{1}{c|}{1} & \multicolumn{1}{c|}{-0.5} & \multicolumn{1}{c|}{1} & \multicolumn{1}{c|}{1} & 3.5 & Multi Pizza & \multicolumn{1}{c|}{0} & \multicolumn{1}{c|}{0} & \multicolumn{1}{c|}{-0.5} & \multicolumn{1}{c|}{0} & \multicolumn{1}{c|}{0} & -0.5 \\
 & &  & \multicolumn{1}{c|}{} & \multicolumn{1}{c|}{} & \multicolumn{1}{c|}{} & \multicolumn{1}{c|}{} & \multicolumn{1}{c|}{} &  & Pizzeria & \multicolumn{1}{c|}{} & \multicolumn{1}{c|}{} & \multicolumn{1}{c|}{} & \multicolumn{1}{c|}{} & \multicolumn{1}{c|}{} &  \\ \hline
\end{tabular}
\end{table}

The results of the quantitative analysis are presented in two tables: Table \ref{llms_Agree} provides the evaluation of the names on which both LLMs agree and Table \ref{llms_Disagree}, where the LLMs choose a different name. 

\paragraph{Rating agreed names} 
There were 16 concepts out of 28 on which both LLMs agree on their first choices.  
To assess the concordance between human judgment and LLM  choices, we selected five persons (\texttt{J1} to \texttt{J5} in the tables), skilled in modeling and terminology, and familiar with the data domain. 
The assessment rules are the following: (i) if the human judge agrees with the name proposed by the LLM the grade is $1$; (ii) if the human judge prefers another name to the LLMs main choice, that name being in the union of the list of proposed names by both LLMs, then the judge rates the choice $0.5$ and  writes  the preferred name; (iii) if the human judge  approves neither the first choice name, nor the ones in the proposal lists union and do not suggest any other name of their own, then the LLM first choice is rated $0$; (iv) last, if the LLM first choice is not accepted but the judge has another choice of their own, then the LLM first choice is rated $-0.5$. 
The rationale is the following: to what extent  is there a consensus between human choices and LLMs selection when both LLMs agree? So the closer to $5$ the score, the more acceptable the “artificial agent" proposal is. 
When looking at Table \ref{llms_Agree}, 11 out of 16 concepts rate a full $5$ score, meaning that there is a total agreement about the name. On the 5 remaining concepts, let us note that the scores are quite low, between $0$ and $-1.5$. This indicates that human judges have  either better choices than those suggested by the LLMs (see previous qualitative discussion)  or are not at ease with the data model itself. Moreover, on these concepts, it seems that there is a concordance between judges \texttt{J1} to \texttt{J4}, \texttt{J5} being apart. When looking at the names, the 4 first judges prefer precise and non-ambiguous names even if they are longer, whereas the last one stood by a more contextual, shorter name. The variation is interesting because human judgment is generally broad and shows variety, and this is reflected in our judges panel. 
For concepts \texttt{CPizzer33} and \texttt{CPizzer34}, the first four judges suggest very different names: “red-pizza serving pizzeria", “BBQ-pizza serving pizzeria"  or the verbal pattern “pizzeria serving red pizza" (respectively BBQ pizza). For concepts \texttt{CPizzer35} and \texttt{CPizzer36}, those they focus on most, they also provide a similar structure, which is sometimes found in the LLM lists of proposals.  

So when the LLMs agree, there are two cases: either the chosen name is adopted by human judges, or it is quite strongly rejected (no middle value scores). 

\paragraph{Rating disagreed names}
In Table \ref{llms_Disagree}, 12 out of 28 names selected by the LLMs present different first choices. 
The judges evaluate here the most fitting LLM and their judgment is differential. The evaluation rules are the following: (i) if name A is rated 1 (accepted) then name B must be rated 0; (ii) Names A and B are rated $0.5$  if there is a better name in the union of the LLMs lists of proposals. Both have the same value if they are equidistant, one of them at $0$ if it is less good than the other; (iii) Both names are rated 0 if they are not accepted and the judge has no proposition; (iv) Names are rated $-0.5$ if they are not accepted and there is a proposition outside the lists, either both, or at least one of them, the “worse" one, the other scores 0. 

The rationale of these rules is the following: whether one of the LLMs outperforms the other and if so, to what extent and in which cases? 
\\ A first answer is to sum the scores of the 12 concepts for each LLM. Claude adds up to 38 whereas ChatGPT obtains 5. There is a discriminant judgement in favor of Claude choices. 
\\ A second track for analysis is to locally examine the scores. A score between 4 and 5 is considered as a global agreement. In Table \ref{llms_Disagree}, 5 names from Claude have obtained a value superior or equal to 4 whereas ChatGPT presents none. 
In contrast, Claude names rating less than 2 are only 2 against 10 for ChatGPT. The trend sways distinctly in favor of Claude. 
Middle values, rating from $2.5$ to $3.5$ show a wide distribution among judges. Here difference is the most noticeable. 
\\Two concept names chosen by ChatGPT have better grades than the corresponding Claude ones: it is for the “ingredient" top concept \texttt{CIng9} (respectively 0 for Claude and 2 for ChatGPT) and concept \texttt{CPizza23} with respectively 0 and $1.5$.  This could be an indication that the data model lacks enough information. 
According to this differential assessment, it seems that Claude is closer to the judges' panel overall opinion.

\paragraph{Summarizing the quantitative analysis}
Naming agreement between the LLMs appears on $57.14\%$ of the total concepts ($16/28$). Out of this, total agreement between human judges and LLMs rates $68.75\%$ of the preceding percentage ($11/16$), which makes an absolute agreement between LLMs and judges (on all the concept names) of $39.28\%$ ($11/28$). 
\\Naming disagreement, on the remaining $42.86\%$ ($12/28$), is the part that helps distinguish between the LLMs. There is a global agreement of the human judges with Claude on about $41.66\%$ of its proposed names ($5/12$), and there is no agreement at all with ChatGPT. 
\\So as a whole, Claude first choice names fit in 16 concepts over 28 ($57.14\%$), in common or differential naming tables,  whereas ChatGPT choice fits only in the common table ($39.28\%$, or $11/28$). Claude is more often aligned with the judges in this example. 
\\Considering the variation scope in judges rating: naming a concept depends on what each individually favors and the judge's personal point of view. If judges favor absolute accuracy (with a bit of redundancy) then the proposed name is longer, and they sometimes produce a verbal phrase pattern instead of a noun phrase pattern. If they prefer a trade-off between redundancy and elision, accepting a contextual more or less elliptic naming (since context disambiguates), desiring to stick to the data model in its present state, then items of agreement between both LLMs would score well in their opinion, and the list of name proposals is a helpful tool for their choice. In many cases, Claude wins, but ChatGPT is sometimes more to the point, not necessarily in its first choice, but in its list of proposals. 

When human judges are domain experts, they can also assess the quality of the underlying data model by examining the accuracy of the proposed concept names, then the deviation between their own ideas and the LLMs results might help them consider revisiting the entire data model. Negative and null values that appear a lot tend to show that the data of the model is either too scarce, or ambivalent. Pizzeria  concepts from 33 to 36, Ingredient concept 9 (top) and pizza concept 23 might need a thorough consideration. As a first indication, their issue seems to incline towards data scarcity for the pizzeria and pizza concepts and a gap in the model for the ingredient. A second thought would be  that what is not easily named might be ill-defined. This is an interesting side effect of the naming procedure.

\section{Related Work}
\label{sec_related}

This section reviews related work from the broader context to the specific setting of FCA and RCA. We first discuss LLM-assisted knowledge and ontology engineering, which motivates the use of LLMs for semantic formulation and knowledge-representation tasks (Sect.~\ref{sec_rwllmass}). We then consider work on labeling generated abstractions, such as clusters and topics, whose objective is to make automatically produced structures interpretable by users (Sect.~\ref{sec_rwlab}). Finally, we review concept naming in FCA and RCA, the closest line of work to the problem addressed here (Sect.~\ref{sec_rwfca}).

\subsection{LLM-assisted knowledge and ontology engineering}
\label{sec_rwllmass}

Recent work has investigated the use of large language models in ontology and knowledge engineering. Li et al. provide a systematic literature review of LLM-based approaches for ontology engineering, covering activities such as requirements specification, ontology implementation, documentation, maintenance, and evaluation \cite{li2026llm4oe}. Garijo et al. further structure this emerging area by identifying ontology-engineering tasks that may benefit from LLMs, while emphasizing the need for clearer benchmarks and evaluation protocols \cite{garijo_llm}. More broadly, Shimizu and Hitzler discuss how LLMs may accelerate knowledge graph and ontology engineering tasks, including ontology modeling, extension, modification, population, alignment, and entity disambiguation \cite{b53a953766154a96b6963154a546150c}.

Our work is complementary to these approaches. We do not use an LLM to directly build, extend, align, populate, or repair an ontology or a knowledge graph. Instead, we use it as a terminological assistant for naming symbolic abstractions generated by FCA or RCA. The LLM input is controlled by a variability model that specifies which formal information is made available for naming, so that naming suggestions can be compared across explicit configurations.

\subsection{Labeling generated abstractions}
\label{sec_rwlab}

A related line of work concerns the automatic labeling or description of generated abstractions, such as clusters and topics. Glover et al. address the problem of inferring hierarchical descriptions for a topic from a small set of example Web pages. Their model distinguishes between self terms, which describe the cluster itself and can be used as candidate names, parent terms, which describe more general concepts, and child terms, which describe specializations of the cluster \cite{10.1145/584792.584876}. This distinction is relevant to our setting, since naming a generated unit may depend not only on its own descriptors, but also on its position with respect to more general and more specific units.

Hierarchical cluster labeling more directly addresses the assignment of meaningful descriptors to nodes of a generated hierarchy. Treeratpituk and Callan propose an algorithm that evaluates candidate labels using information from the cluster, the parent cluster, and corpus statistics, with the aim of producing labels that support browsing and interpretation of hierarchical organizations \cite{10.1145/1146598.1146650}. Similarly, topic labeling aims at producing concise labels for latent topics learned from text collections. Lau et al. generate candidate labels from top-ranking topic terms, Wikipedia article titles, and sub-phrases, and then rank these candidates using association measures and lexical features \cite{10.5555/2002472.2002658}.

These works share with ours the objective of turning generated structures into human-interpretable units. However, FCA and RCA concepts have a different status: they are symbolic abstractions formally defined by their intent, extent, order relations, implications, and, in RCA, relational attributes. Concept naming therefore has to account not only for local descriptors, but also for the position of a concept in a structured space and for dependencies with related concepts. This motivates a configurable framework that makes explicit which parts of the formal structure are used to guide naming. Another difference is methodological. Most of these approaches rely on the extraction, scoring, and ranking of candidate terms using statistical, lexical, or external-resource-based features. In contrast, we use an LLM as a terminological assistant able to formulate candidate names from structured symbolic descriptions. The LLM is not used in isolation, however: its input is controlled by a variability model that specifies which formal information is exposed during naming.

\subsection{Concept naming in FCA and RCA}
\label{sec_rwfca}

Works in the FCA domain considering terminology are quite present, with most of them focusing on applying FCA techniques to lexical data. 
Authors have explored  terminological and linguistic applications of FCA \cite {DBLP:conf/fca/Priss05,DBLP:phd/dnb/Priss98},  creation or expansion of ontological terminologies and thesauri \cite{9313186},  information retrieval \cite{DBLP:journals/amai/CodocedoLN14,Carpineto1996},
translation \cite{DBLP:conf/icfca/PrissO09} \cite{DBLP:conf/iccs/PrissO07}, terminological weeding \cite{DBLP:conf/kont/PrissO07}, and semantic annotation tasks \cite{Cigarran2014SemanticAnnotationFCA}. 
The dataset was mostly a lexical database, a thesaurus, a social network \cite{Missaoui2017TerminologyFCA}, a terminologically structured data, or a conceptual lexical database  
\cite{PrissOld2004}. 
In some studies, corpora of general texts have also been used to construct ontological structures, 
or to discover concepts and semantic relations \cite{IgnatovKuznetsov2013FCAIR}. 
Terminological issues in FCA are also indirectly connected to \cite{Formica2021ConceptSimilarityFCA}, which investigates concept similarity in many-valued contexts and thereby contributes to the semantic comparison of concepts. 
Conversely, in more recent studies, the terminology defines the process and the FCA or RCA model provides the dataset.
They are part of a line of research that includes
 automatically generating textual explanations of recurring lattice structures  \cite{HirthHornStummeHanika2023TextualExplanations}, and  addressing the concept naming issue in \cite{arandaConcepts2024,aranda_corral_2026_18608706,DBLP:conf/concepts/GuenouneGHLMMZ25}, that we develop below.

In RCA, developing the relational attributes as performed in \cite{DBLP:journals/ijgs/DolquesBHG16,DBLP:conf/icfca/DolquesBHB19,DBLP:conf/kcap/WajnbergVLPM19,DBLP:conf/concepts/MusslinBHMPRS24,DBLP:conf/concepts/GutierrezHMZ25} is intended to facilitate understanding of the implications and concepts.
This is critical for implications from RCA, because they typically contain relational attributes that refer to concepts. But when relational attributes are lengthy due to the complexity of the intentions, this approach reaches its limits: naming concepts and using their name may 
be a relevant alternative or a complementary solution.

Moreover, De Maio et al. \cite{DBLP:journals/apin/MaioFGLS14} propose a related approach in the context of fuzzy FCA/RCA-based semantic annotation. Although their primary objective is ontology construction rather than concept naming, the generated ontology classes are assigned labels through a function that selects the most representative datatype properties. This makes their work relevant to ours, but the perspective remains different: in their case, naming is a component of ontology generation, whereas in our work, it is treated as a central interpretive and terminological task.

The most related works are quite recent and thus rely on large language models (LLMs) as a terminological resource and expert (following \cite{10.1145/3597503.3639180}) to provide lexical labels to concepts. 
Aranda et al. \cite{arandaConcepts2024,aranda_corral_2026_18608706} assess the issue of naming concepts that do not introduce any attribute in a concept lattice, with a strategic preference based on pruning information from siblings, and relying on attributes. 
They propose an incremental approach to concept naming that consists in exploiting the structure of the concept lattice itself to guide the naming process. In this setting, attribute introducing concepts are first associated with names derived from their declared attributes, whereas the other concepts remain anonymous. Anonymous concepts are then processed when all their direct parents have already been assigned a name, thus enabling a top-down progression from abstract to more specific concepts. 
At each iteration, a prompt is constructed from several information sources, including a general scope term, direct parent attributes, examples drawn from the extent, and contrasting objects absent from the target concept took from sibling concepts. The LLM is then asked to produce a candidate term, which may in turn be reintroduced into the structure as a new attribute attached to the corresponding concept. Such approach views concept naming as a progressive and context-dependent process, in which previously assigned names contribute to the interpretation of subsequent concepts.
While promising, this approach also exhibits several limitations. It is mainly designed to produce attribute-like labels rather than broader group names, which restricts the nature of the naming output and raises the question of whether concept names should be treated as attributes only. It may further become difficult to use when concepts accumulate many introduced attributes, since the resulting prompts may grow and become too complex to yield concise and intelligible names. Moreover, requiring a single term only is limiting, as several candidate names may be relevant depending on the expected degree of precision or readability. The contrasting information considered in the prompt is also partial, as it is restricted to objects. In addition, object introducing concepts, especially those whose extent is reduced to a single object, may require a dedicated treatment and could in some cases be regarded as non-anonymous. Nevertheless, this line of work is a valuable source of inspiration for the present study.

Concept naming has also been proposed  in the context of RCA-based UML model refactoring \cite{DBLP:conf/concepts/GuenouneGHLMMZ25}. In this setting, the RCA framework relies on formal contexts describing UML classes, attributes, operations, and roles, connected through relational contexts. The naming task is coupled with a relevance assessment, with the objective of identifying new superclasses that could improve the factorization and structuring of the model by making new abstractions emerge.
Compared with the approach of Aranda et al. \cite{arandaConcepts2024,aranda_corral_2026_18608706}, in which names may be propagated through the conceptual structure and reintroduced as attributes, the generated name is not used here as a new attribute, but rather as a candidate class name for the UML model itself. Only class concepts that are not introducers of a single object are named, whereas concepts introducing a single object (a UML class) inherit the name of this corresponding UML class. In addition, all attributes are provided to the LLM so that each concept description remains
self-contained, although inherited elements are explicitly marked to preserve the distinction between general and concept-specific information. Finally, unlike approaches that exploit the lattice hierarchy to guide naming, concepts are processed in an arbitrary order on the basis of their description alone.
This approach also has several limitations. First, it does not exploit any contrasted part, whether attributes or objects, that could help distinguish a concept from closely related ones. Second, all attributes are systematically included in the description, which may result in long and complex prompts. This may partly compensate for the absence of contrasted information, but it also increases the cognitive load of the naming task. Third, all attributes are treated in the same way, although deeply nested relational attributes may introduce unnecessary complexity and emphasize secondary information, which becomes particularly problematic in large UML models, unless the RCA process is intentionally stopped after only a few steps. Finally, it is arguably a missed opportunity not to exploit the hierarchy of the lattice more directly in the naming process. Despite these limitations, this use case shows how RCA-based analysis and LLM-assisted naming can be articulated with model restructuring objectives.

In this framework, the present paper builds on an integrates the contributions of both previous ones~\cite{DBLP:conf/concepts/GuenouneGHLMMZ25,DBLP:conf/concepts/GutierrezHMZ25}, but with a special highlight on terminological and variability-based prompt generation strategies, in order to improve naming quality and automating as much as possible the name selection process.

\section{Conclusion}
\label{sec_conclusion}

The study presented in this paper shows that concept naming in the context of Relational Concept Analysis is not a trivial post-processing step, but a complex interpretive task shaped by linguistic, structural, and methodological factors.
To address this task, we proposed a variability-based framework for LLM-assisted concept naming. The framework combines the generative capabilities of LLMs with a product-line approach, making it possible to explicitly model and explore different naming configurations. These configurations account for several dimensions: the linguistic properties of natural language, information available in the concept structures, the way this information is represented in prompts, and the preferences or constraints specified by the analyst.
We illustrated the approach on a small relational dataset.

The results have shown that ChatGPT 5.2 Thinking and Claude Sonnet 4.6 Extended web applications tend to agree and produce humanly acceptable names when concepts are supported by clear attributes and relations. By contrast, naming becomes much more variable for less well-characterized concepts, revealing different preferences in the LLMs.
Although the illustration was made with a limited set of concepts extracted from a small and simple dataset, it provides initial evidence that prompt-based support for concept naming can be systematically organized and studied.
The size and nature of the  illustration dataset were decided as reasonable features for a "proof of concept". Size may introduce new phenomena that have to be dealt with as related to scope, as explained in the introduction. The nature of the dataset could be either very technical or general knowledge. Domain-specific data is not easily understandable for a reader, whereas general knowledge could be more accessible. 
The reader may thus appreciate the contribution of this naming process. 
More broadly, prompt-based support for concept naming opens the way to a variability-aware view of LLM-assisted naming, in which prompt generation is not ad hoc, but guided by explicit choices regarding the information to be mobilized and the expected style of the naming outcome.

Future work will address both tool and methodology-related aspects, as well as issues related to dataset analysis, terminology, and the role played by LLMs in the overall process.
A first research direction is to refine the modeling of variability in the prompt generation process, with the aim of identifying more precisely which combinations of information best support concept naming in different situations. 
We will further consolidate the proposed Software Product Line approach to prompt generation and concept naming. Although much of the variability-aware pipeline is already in place, it could be improved in several directions, including the systematic generation and writing of prompt templates, stronger tool support for configuration and input preparation through both LLMs and dedicated programs, and richer forms of analyst-facing output for presenting selected names together with relevant alternatives. About concept naming, other name patterns have been chosen by human judges, and thus the name pattern  set could be enriched: the noun phrase pattern for names could be completed with a present participle verb phrase pattern that would broaden the perspective and favor understanding. 

A particularly promising direction for the analysis of RCA datasets is to benefit from the iterative nature of RCA. 
In its current form, the approach operates on concepts observed at a given step of the RCA process, and at the final step in the present study. In the longer term, it could support an incremental naming process, e.g. with RCAviz\footnote{https://rcaviz.lirmm.fr/}, including revision and stabilization, as conceptual structures are progressively refined.

It would also be of interest to investigate whether LLMs, while analyzing and naming concepts, can help reveal terminological weaknesses in the input dataset by highlighting ambiguities and missing information they encounter. 
They could then help address these weaknesses, for example by suggesting more precise terms, additional attributes, a clearer categorization of attributes, or additional formal and relational contexts. These additional contexts would explicitly distinguish secondary parts of the descriptions, expressed through relations, from more primary ones captured through the primitive attributes.

An important issue concerns the status of the relations in RCA datasets. In RCA, relations are provided by the analyst. As they are not subject to generalization, they may impede the naming process. Thus a relevant question for future work is whether this approach impacts concept naming.
To support the generalization of, and reasoning about, relations themselves, a likely way forward would be to investigate a modeling strategy similar to the one developed in RCA-based approaches for the refactoring of UML class models. In the latter, the relational context family respects the structure of the UML metamodel, giving relations the status of objects themselves described by attributes and relations.

Finally, this work will be extended to datasets that are more diverse in terms of size, domain of application, and structural complexity. It will particularly consider datasets such as UML models to be refactored \cite{VERNIER201311}, datasets from the Knomana initiative \cite{plants2021}, as well as datasets from the hydroecological domain \cite{DBLP:conf/iccs/NicaBB18}, and ancient pharmacopoeia \cite{DBLP:conf/icfca/BraudDFLBP21}. A main benefit is that these datasets have already been studied using FCA and RCA, and domain experts are available to carry out qualitative evaluations in collaboration with a terminology expert.


\section*{Declaration on the use of generative AI}

Beyond language editing, generative AI tools, including ChatGPT, were used during manuscript preparation to support section structuring, reformulation, positioning, and literature exploration. AI-generated suggestions, especially concerning related work, were carefully checked, completed, and corrected by the authors. Generative AI tools were not used to produce the scientific results or make final scholarly decisions. All AI-assisted material was reviewed, edited, and validated by the authors, who remain fully responsible for the content of the manuscript.


\paragraph{Acknowledgement}
This work was supported by ANR SmartFCA, Grant ANR-21-CE23-0023 of the French National Research Agency. The authors also warmly thank Gonzalo A. Aranda-Corral for fruitful discussions during the preparation of this work.

\bibliography{biblioshort}

\end{document}